\documentclass{article}

\PassOptionsToPackage{numbers, compress}{natbib}

\usepackage[preprint]{neurips_2023}

\usepackage{tcolorbox}
\definecolor{cadmiumgreen}{rgb}{0.0, 0.42, 0.24}
\definecolor{oldmauve}{rgb}{0.4, 0.19, 0.28}
\definecolor{royalazure}{rgb}{0.0, 0.22, 0.66}
\definecolor{harvardcrimson}{rgb}{0.79, 0.0, 0.09}
\definecolor{lightmauve}{rgb}{0.86, 0.82, 1.0}
\definecolor{darkbrown}{rgb}{0.4, 0.26, 0.13}%
\usepackage[colorlinks = true,
            linkcolor = royalazure,
            urlcolor  = royalazure,
            citecolor = royalazure,
            anchorcolor = royalazure]{hyperref}
            
\usepackage{amsmath,amsfonts,bm}

\def\1{\bm{1}}

\def\vtheta{{\bm{\theta}}}

\def\vg{{\bm{g}}}

\def\vp{{\bm{p}}}

\def\vx{{\bm{x}}}

\def\mV{{\bm{V}}}

\def\mY{{\bm{Y}}}

\DeclareMathAlphabet{\mathsfit}{\encodingdefault}{\sfdefault}{m}{sl}
\SetMathAlphabet{\mathsfit}{bold}{\encodingdefault}{\sfdefault}{bx}{n}

\def\gA{{\mathcal{A}}}

\def\gD{{\mathcal{D}}}

\def\gR{{\mathcal{R}}}
\def\gS{{\mathcal{S}}}

\def\gV{{\mathcal{V}}}

\def\gX{{\mathcal{X}}}
\def\gY{{\mathcal{Y}}}

\DeclareMathOperator*{\argmax}{arg\,max}

\usepackage[utf8]{inputenc} 
\usepackage[T1]{fontenc}    
\usepackage{hyperref}       
\usepackage{url}            
\usepackage{booktabs}       
\usepackage{amsfonts}       
\usepackage{amsthm}
\usepackage{nicefrac}       
\usepackage{microtype}      
\usepackage{xcolor}         

\usepackage{algorithm}
\usepackage{algorithmic}
\usepackage{float}
\usepackage{booktabs}
\usepackage{multirow}
\usepackage{capt-of}
\usepackage{wrapfig,lipsum,booktabs}
\usepackage{enumitem,graphicx}

\title{Shared Adversarial Unlearning: Backdoor Mitigation by Unlearning Shared Adversarial Examples}

\newcommand*{\email}[1]{%
    \normalsize\href{mailto:#1}{#1}\par
    }
\author{%
\ \ \ \ Shaokui Wei\textsuperscript{1,2}
\ \ \ \ Mingda Zhang\textsuperscript{1,2}
\ \ \ \ Hongyuan Zha\textsuperscript{1}
\ \ \ \ Baoyuan Wu\textsuperscript{1,2}\thanks{Corresponds to Baoyuan Wu (\email{wubaoyuan@cuhk.edu.cn}).} 
\\
\textsuperscript{1}School of Data Science, \\The Chinese University of Hong Kong, Shenzhen (CUHK-Shenzhen), China\\
\textsuperscript{2}Shenzhen Research Institute of Big Data}

\def\Db{\mathcal{D}_{cl}}
\def\Ds{\mathcal{D}_{s}}

\newcommand{\RNum}[1]
    {\MakeUppercase{\romannumeral #1}}
\def\htt{h_\vtheta}
\def\hbd{h_{\vtheta_{bd}}}
\def\Dn{\gD_{-\hat{y}}}

\def\app{\textbf{Appendix}}

\def\ind{\mathbb{I}}
\def\veps{\bm{\epsilon}}

\newtheorem{proposition}{Proposition}
\newtheorem{assumption}{Assumption}
\newtheorem{definition}{Definition}

\begin{document}

\def\ie{$i.e.$}
\def\eg{$e.g.$}
\def\etc{$etc$}
\newcommand{\wrt}{\textit{w.r.t.~}}
\long\def\comment#1{}
\def\red#1{\textcolor{red}{#1}}
\def\blue#1{\textcolor{blue}{#1}}
\def\revise#1{\textcolor{blue}{#1}}
\newtheorem{appass}{Assumption}[subsection]
\newtheorem{appprop}{Proposition}[subsection]
\maketitle

\begin{abstract}
Backdoor attacks are serious security threats to machine learning models where an adversary can inject poisoned samples into the training set, causing a backdoored model which predicts poisoned samples with particular triggers to particular target classes, while behaving normally on benign samples. In this paper, we explore the task of purifying a backdoored model using a small clean dataset. By establishing the connection between backdoor risk and adversarial risk, we derive a novel upper bound for backdoor risk, which mainly captures the risk on the shared adversarial examples (SAEs) between the backdoored model and the purified model. This upper bound further suggests a novel bi-level optimization problem for mitigating backdoor using adversarial training techniques. To solve it, we propose \textbf{S}hared \textbf{A}dversarial \textbf{U}nlearning (SAU). Specifically, SAU first generates SAEs, and then, unlearns the generated SAEs such that they are either correctly classified by the purified model and/or differently classified by the two models, such that the backdoor effect in the backdoored model will be mitigated in the purified model. Experiments on various benchmark datasets and network architectures show that our proposed method achieves state-of-the-art performance for backdoor defense. 

\end{abstract}

\section{Introduction}
Recent decades have witnessed a growing application of deep neural network (DNN) techniques in various domains, such as face recognition, autonomous driving, and medical image processing \cite{adjabi2020past, he2016deep, liu2020computing, tournier2019mrtrix3, xiao2018iot}. However, DNNs are known to be vulnerable to malicious attacks that can compromise their security and reliability. One of the emerging threats  is the backdoor attack, where an attacker attempts to inject stealthy backdoor into a DNN model by manipulating a small portion of the training data such that the poisoned model will behave normally for clean inputs while misclassifying any input with the trigger pattern to a target label. Since the poisoned models can have real-world consequences, such as allowing an attacker to gain unauthorized access to a system or to cause physical harm, it is crucial to develop effective methods for backdoor defense in developing and deploying DNNs.

 Adversarial training (AT) is one of the most effective methods for improving the robustness of DNNs \cite{madrytowards}.  AT is typically formulated as a min-max optimization problem, where the inner maximization aims to find adversarial perturbations that fool the model, while the outer minimization is to reduce the adversarial risk. 
 Recently, several works have explored the use of AT for defending against backdoor attacks \cite{tradeoff_backdoor,zeng2022adversarial}. However, these works face some limitations and challenges. For example, \citet{tradeoff_backdoor} found that increasing the robustness to adversarial examples may increase the vulnerability to backdoor attacks when training from scratch on poisoned datasets. \citet{zeng2022adversarial} proposed to unlearn the universal adversarial perturbation (UAP) to remove backdoors from poisoned models. Since directly unlearning UAP leads to highly unstable performance for mitigating backdoors, they employ implicit hyper gradient to solve the min-max problem and achieve remarkable performance. However, their method assumes that the same trigger can activate the backdoor regardless of the samples it is planted on, and therefore lacks a guarantee against more advanced attacks that use sample-specific and/or non-additive triggers.
 
In this paper, we consider the problem of purifying a poisoned model. After investigating the relationship between adversarial examples and poisoned samples,  a new upper bound for backdoor risk to the fine-tuned model is proposed. Specifically, by categorizing adversarial examples into three types, we show that the backdoor risk mainly depends on the shared adversarial examples that mislead both the fine-tuned model and the poisoned model to the same class. Shared adversarial examples suggest a novel upper bound for backdoor risk, which combines the shared adversarial risk and vanilla adversarial risk. Besides, the proposed bound can be extended with minor modifications to universal adversarial perturbation and targeted adversarial perturbation. Based on the new bound, we propose a bi-level formulation to fine-tune the poisoned model and mitigate the backdoor. To solve the bi-level problem, we proposed  \textbf{S}hared \textbf{A}dversarial \textbf{U}nlearning (SAU). SAU first identifies the adversarial examples shared by the poisoned model and the fine-tuned model. Then, to break the connection between poisoned samples and the target label, the shared adversarial examples are unlearned such that they are either classified  correctly by the fine-tuned model or differently by the two models. Moreover, our method can be naturally extended to defend against backdoor attacks with multiple triggers and/or multiple targets. To evaluate our method, we compare it with six state-of-the-art (SOTA) defense methods on seven SOTA backdoor attacks with different model structures and datasets. Experimental results show that our method achieves comparable and even superior performance to all the baselines.

Our contributions are three folds: \textbf{1)} We analyze the relationship between adversarial examples and poisoned samples, and derive a novel upper bound for the backdoor risk that can be generalized to various adversarial training-based methods for backdoor defense; \textbf{2)} We formulate a bi-level optimization problem for mitigating backdoor attacks in poisoned models based on the derived bound, and propose an efficient method to solve it; \textbf{3)} We conduct extensive experiments to evaluate the effectiveness of our method, and compare it with six state-of-the-art defense methods on seven challenging backdoor attacks with different model structures and datasets.

\section{Related work}
\paragraph{Backdoor attack}

Backdoor attack is one of the major challenges to the security of DNNs. The poisoned model behaves normally on clean inputs but produces the target output when the trigger pattern is present \cite{wu2023adversarial}.  Based on the types of triggers,   Backdoor attacks can be classified into two types based on the triggers: fixed-pattern backdoor attacks and sample-specific backdoor attacks. BadNets \cite{gu2019badnets} is the first backdoor attack that uses fixed corner white blocks as triggers. To improve the stealth of the triggers, Blended \cite{chen2017targeted} is proposed to blend the trigger with the image in a weighted way. Since fixed-pattern triggers can be recognized easily, sample-specific backdoor attacks have been proposed. SSBA \cite{li2021invisible}, WaNet \cite{nguyen2021wanet}, LF \cite{zeng2021rethinking} use different techniques to inject unique triggers for different samples from different angles. Sleeper-agent \cite{souri2022sleeper} and Lira \cite{doan2021lira} optimize the target output to obtain more subtle triggers. Recently, \cite{zi2023boost} propose a learnable poisoning sample selection strategy to further boost the effect of backdoor attack. To keep the label of the backdoor image matching the image content, LC \cite{shafahi2018poison} and SIG \cite{barni2019new}  use counterfactual and other methods to modify the image to deploy a clean label attack.

\paragraph{Backdoor defense}

Backdoor defense aims to reduce the impact of backdoor attacks on deep neural networks (DNNs) through training and other means. There are two types of backdoor defense: post-processing and in-processing. Post-processing backdoor defense mitigates the effect of backdoors for a poisoned model by pruning or fine-tuning.  For example, FP \cite{liu2018fine} prunes some potential backdoor neurons and fine-tunes the model to eliminate the backdoor effect; ANP \cite{wu2021adversarial} finds the backdoor neurons by adversarial perturbation to model weights; EP \cite{zheng2022preactivation} distinguishes the statistical performance of the backdoor neurons from the clean neurons; NAD \cite{li2021neural} uses a mild poisoned model to guide the training of the poisoned model to obtain a cleaner model; I-BAU \cite{zeng2022adversarial} finds possible backdoor triggers by the universal adversarial attack and unlearn these triggers to purify the model. In-processing backdoor defense reduces the effect of the backdoor during training. For instance, ABL \cite{li2021anti} exploits the fact that the learning speed of backdoor samples is faster than that of the clean sample, therefore they split some poisoned image and eliminates the backdoor by forgetting these backdoor samples; DBD \cite{huang2022backdoor} divides the backdoor training process and directly inhibits the backdoor learning process; D-ST \cite{chen2022effective} splits the backdoor samples and uses semi-supervised learning by observing that the clean samples are more robust to image transformation than the backdoor samples.

\paragraph{Adversarial training}

In adversarial training, models are imposed to learn the adversarial examples in the training stage and therefore, resistant to adversarial attacks in the inference stage. In one of the earliest works \cite{goodfellow2014explaining}, the adversarial examples are generated using Fast Gradient Sign Method. In \cite{madrytowards}, PGD-AT is proposed, which generates adversarial examples by running FGSM multiple steps with projection and has become the most widely used baseline for adversarial training. Some further improvements of PGD-AT include initialization improvement \cite{jia2022prior, jia2022boosting}, attack strategy improvement \cite{jia2022adversarial}, and efficiency improvement \cite{wongfast, zheng2020efficient}. 

\section{Methodology}
In Section~\ref{sec::pre}, we first introduce notations, threat model, and defense goal to formulate the problem. By investigating the relationship between adversarial risk and backdoor risk, a new upper bound of backdoor risk is derived in Section~\ref{sec::risk}, from which a bi-level formulation is proposed in Section \ref{sec::loss}.

\subsection{Preliminary}
\label{sec::pre}

\paragraph{Notations.} 
Let uppercase calligraphic symbols denote sets/spaces and risks. We consider a $K$-class ($K\geq 2$) classification problem that aims to predict the label $y\in \gY$ of a given sample $\vx \in \gX$, where $\gY=[1, \cdots, K]$ is the set of labels and $\gX$ is the space of samples. Let $h_{\vtheta}:\gX\to \gY$ be a DNN classifier with model parameter $\vtheta$. We use $\gD$ to denote the set of data $(\vx,y)$.
For simplicity, we use $\vx\in \gD$ as a abbreviation for $(\vx,y)\in \gD$. Then, for a sample $\vx\in \gD$,  its predicted label is
\begin{equation}
    \htt(\vx) = \argmax_{k=1,\cdots, K} \vp_k(\vx;\vtheta), 
\end{equation}
where $\vp_k(\vx;\vtheta)$ is the (softmax) probability of $\vx$ belonging to class $k$. 

\paragraph{Threat model.}
Let $\gV$ be the set of triggers and define $g:\gX\times \gV \to \gX$ as the generating function for poisoned samples. Then, given a trigger $\Delta \in \gV$ and a sample $\vx\in \gX$, one can generate a poisoned sample $g(\vx;\Delta)$. For simplicity, we only consider all to one case, \ie, there is only one trigger $\Delta$ and one target label $\hat{y}$. The all to all case and multi-trigger case are left in the \app. We assume that the attacker has access to manipulate the dataset and/or control the training process such that the trained model classifies the samples with pre-defined trigger $\Delta$ to the target labels $\hat{y}$ while classifying clean samples normally.  In addition, we define the poisoning ratio as the proportion of poisoned samples in the training dataset. 

\paragraph{Defense goal.}
We consider a scenario where a defender is given a poisoned  model with parameter $\theta_{bd}$ and a small set of \textit{clean} data $\Db=\{(\vx_i,y_i)\}_{i=1}^N$. Since we are mainly interested in samples whose labels are not the target label, we further define the set of non-target samples as $\Dn = \{(\vx,y) |(\vx,y) \in \Db, y\neq \hat{y}\}$. Note that the size of $\Db$ is small and the defender cannot train a new model from scratch using only $\Db$. The defender's goal is to purify the model so that the clean performance is maintained and the backdoor effect is removed or mitigated. We assume that the defender cannot access the trigger  $\Delta$ or the target label $\hat{y}$.

\paragraph{Problem formulation.}
Using the $0$-$1$ loss \cite{ wang2020improving, zhang2019theoretically}, the classification risk $\gR_{cl}$ and the backdoor risk $\gR_{bd}$ with respect to classifier $\htt$ on $\Db$ can be defined as:
\begin{align}
    \label{eq::risk}
    \gR_{cl}(\htt) = \frac{1}{N}\sum_{i=1}^N \ind(\htt(\vx_i)\neq y_i), \quad
    \gR_{bd}(\htt) = \frac{\sum_{i=1}^N \ind(\htt(g(\vx_i,\Delta))=\hat{y}, \vx_i\in \Dn)}{|\Dn|}
\end{align}
where $\ind$ is the indicator function and $|\cdot|$ denotes the cardinality of a set.

In (\ref{eq::risk}), the classification risk concerns whether a clean sample is  correctly classified, while the backdoor risk measures the risk of classifying a poisoned sample from $\Dn$ to target class $\hat{y}$. 

In this paper, we consider the following problem for purifying a poisoned model:
\begin{equation}
    \label{eq::trade}
    \min_{\vtheta} \gR_{cl}(\htt) + \lambda \gR_{bd}(\htt),
\end{equation}
where $\lambda$ is a hyper-parameter to control the tradeoff between classification risk and backdoor risk.

\subsection{Connection between backdoor risk and adversarial risk}
\label{sec::risk}
Since directly minimizing (\ref{eq::trade}) is impractical, a natural choice is to replace $\gR_{bd}$ with a proper upper bound. Therefore, we delve deeper for adversarial risk as an upper bound of backdoor risk.

Let $\gS$ be a set of perturbations. Given a sample $\vx$ and an adversarial perturbation $\veps$, an adversarial example $\tilde{\vx}_{\veps}$ can be generated by $\tilde{\vx}_{\veps} = \vx+\veps$. We define the set of adversarial example-label pair to $\htt$ generated by perturbation $\veps$ and dataset $\Db$ by
$\gA_{\veps,\vtheta}=\{(\tilde{\vx}_{\veps},y)|\htt(\tilde{\vx}_{\veps})\neq y, (\vx,y)\in \Db\}$ and abbreviate $(\tilde{\vx}_{\veps},y) \in \gA_{\veps,\vtheta}$ by $\tilde{\vx}_{\veps} \in \gA_{\veps,\vtheta}$. Then, the adversarial risk for $\htt$ on $\Dn$ can be defined as
\begin{equation}
    \label{eq::ae_risk}
    \gR_{adv}(\htt) = \frac{\sum_{i=1}^N \max_{\veps_i \in \gS}\ind(\tilde{\vx}_{i, \veps_i} \in \gA_{\veps_i,\vtheta}, \vx_i\in\Dn)}{|\Dn|}.
\end{equation}

To bridge the adversarial examples and poison samples, we provide the following assumption:
\begin{assumption}  
\label{assm:S}
Assume that $g(\vx;\Delta) - \vx \in \gS$ for  $\forall \vx \in \Db$.
\end{assumption}

Assumption \ref{assm:S} ensures that there exists $\veps \in \gS$ such that $\vx+\veps = g(\vx;\Delta)$. As poisoned samples are required to be stealthy and imperceptible to humans \cite{nguyen2021wanet, wubackdoorbench,  zeng2021rethinking}, Assumption \ref{assm:S} is quite mild. For example, an $L_p$ ball with radius $\rho$, \ie, $\gS=\{\veps |\|\veps\|_p\leq \rho \}$, is sufficient to satisfy Assumption \ref{assm:S} with proper $p$ and $\rho$.

Then, we reach the first upper bound of backdoor risk:
\begin{proposition}
    Under Assumption~\ref{assm:S}, $\gR_{adv}$ serves as an upper bound of $\gR_{bd}$, \ie, $\gR_{bd}\leq \gR_{adv}$.
\end{proposition}

\begin{wrapfigure}{r}{0.5\textwidth}
  \begin{center}
    \includegraphics[width=0.48\textwidth]{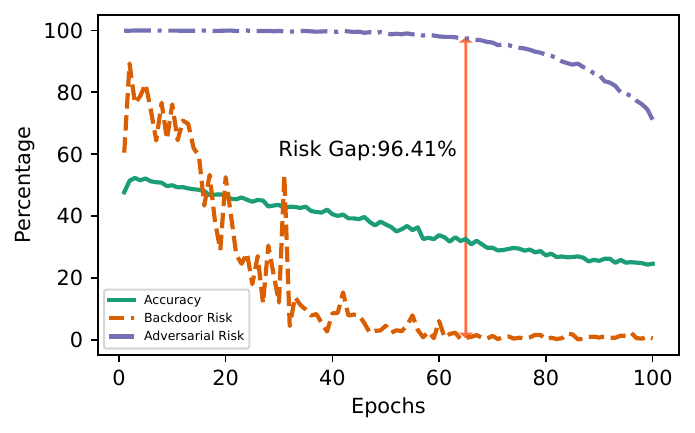}
  \end{center}
  \vspace{-0.2in}
  \caption{Example of purifying poisoned model using adversarial training on Tiny ImageNet  \cite{le2015tiny}. The curves for Accuracy, Backdoor Risk and Adversarial Risk are indicated by Green, Orange and Purple, respectively.}
  \label{gap}
  \vspace{-0.25in}
\end{wrapfigure}

\paragraph{Remark} Although $\gR_{adv}$ seems to be a promising surrogate for $\gR_{bd}$, it neglects some essential connections between adversarial examples and poisoned samples, resulting in a significant gap between adversarial risk and backdoor risk, as shown in Figure \ref{gap}. \emph{Such a gap also leads to a fluctuating learning curve for backdoor risk and a significant  drop in accuracy in the backdoor mitigation process by adversarial training.} Therefore, we seek to construct a tighter  upper bound for $\gR_{bd}$. A key insight is that not all adversarial examples contribute to backdoor mitigation. 

To further identify AEs important for mitigating backdoors, we leverage  the information from the poisoned model to categorize the adversarial examples to $\htt$ into three types, as shown in Table~\ref{tab:AE}.

\begin{table}[h]
  \caption{Different types of adversarial examples to $\htt$. }
  \label{tab:AE}
  \centering
    \scalebox{0.95}{
  \begin{tabular}{lll}
    \toprule
    \cmidrule(r){1-3}
    Type     & Description     & Definition \\
    \midrule
    \RNum{1}      & Mislead $\hbd$ and $\htt$ to the same class       & $\hbd(\tilde{\vx}_{\veps}) = \htt(\tilde{\vx}_{\veps}) \neq y$ \\
    \RNum{2}       & Mislead $\htt$, but not mislead $\hbd$ & $\hbd(\tilde{\vx}_{\veps}) \neq \htt(\tilde{\vx}_{\veps}), \hbd(\tilde{\vx}_{\veps}) = y$     \\
    \RNum{3}       & Mislead $\hbd$ and $\htt$ to different classes  & $\hbd(\tilde{\vx}_{\veps}) \neq \htt(\tilde{\vx}_{\veps}), \hbd(\tilde{\vx}_{\veps}) \neq y, \htt(\tilde{\vx}_{\veps}) \neq y$     \\
    \bottomrule
  \end{tabular}}
\end{table}

\begin{figure}[h]
    \centering
    \includegraphics[width = 0.9\linewidth]{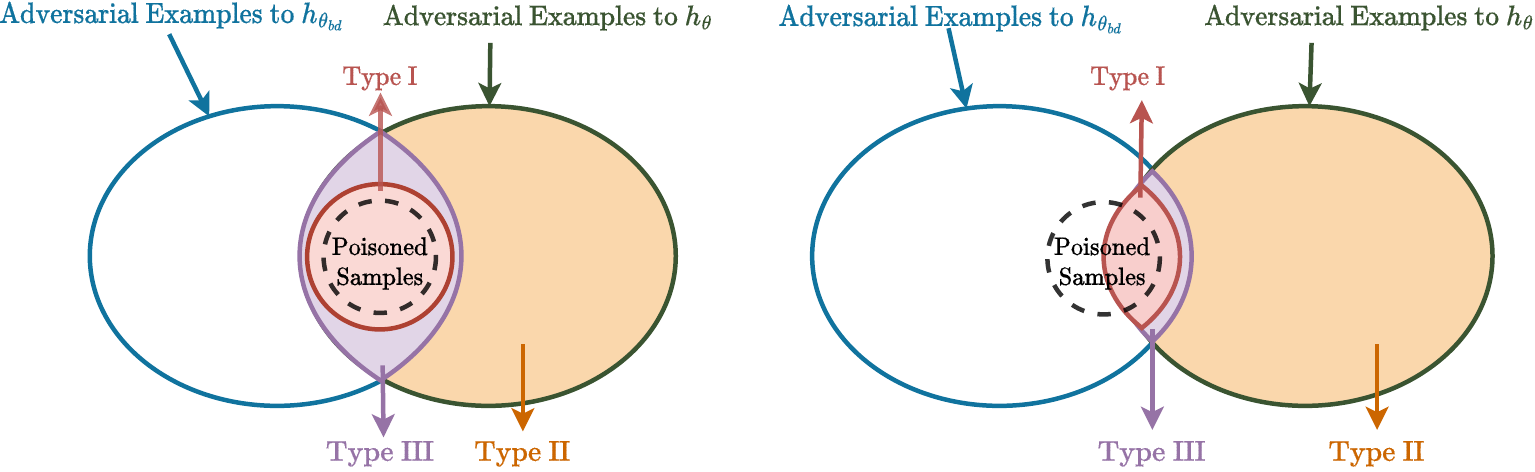}
    \caption{A schematic of the relationship between adversarial examples (SAEs), shared adversarial examples (Type \RNum{1}) and poisoned samples. The adversarial examples for $\hbd$ and $\htt$ are shown in the blue and green solid ellipses, respectively. The poisoned samples are in the black dashed circle. Assume that $\hbd$ and $\htt$ have $100\%$ backdoor attack success rates in the left, such that all poisoned samples are contained in SAEs. Thus, by reducing the shared adversarial examples between $\hbd$ and $\htt$ (from left to right), the backdoor risk can be mitigated in $\htt$.}
    \label{fig::cup}
\end{figure}
Furthermore, we refer adversarial examples of Type \RNum{1} as the shared adversarial examples between $\htt$ and $\hbd$, as defined below:
\begin{definition}[Shared Adversarial Example]
    Given two classifiers $h_{\vtheta_1}$ and $h_{\vtheta_2}$, an adversarial example $\tilde{\vx}_{\veps}$ is shared between  $h_{\vtheta_1}$ and $h_{\vtheta_2}$ if and only if   $h_{\vtheta_1}(\tilde{\vx}_{\veps})=h_{\vtheta_2}(\tilde{\vx}_{\veps})\neq y$.
\end{definition}
Let $\Ds = \{(\vx,y)|\hbd(g(\vx; \Delta))=\hat{y}, \vx\in \Dn\}$ be the subset of samples in $\Dn$ on which planting a trigger can successfully activate the backdoor in $\hbd$. As illustrated in Figure ~\ref{fig::cup}, the following proposition reveals a deeper connection between adversarial examples and poisoned samples.

\begin{proposition} 
\label{ass:share}
For $\forall (\vx,y) \in \Ds$, the poisoned sample $g(\vx; \Delta)$  can attack $\htt$ if and only if 
$\htt(g(\vx; \Delta))=\hbd(g(\vx; \Delta))\neq y.$ Furthermore,  under Assumption \ref{assm:S},  the poisoned sample $g(\vx; \Delta)$ is a shared adversarial example between $\htt$ and $\hbd$, if it can attack $\htt$.
\end{proposition}

Then, the adversarial risk $\gR_{adv}$ can be decomposed to three components as below:
\begin{equation}
\label{eq::decom}
\begin{aligned}
    &\gR_{adv}(\htt) = \frac{1}{|\Dn|} \sum_{i=1}^N \max_{\veps_i \in \gS} \Big\{\underbrace{\ind(\vx_i\in \Ds)\ind(\htt(\tilde{\vx}_{i, \veps_i})=\hbd(\tilde{\vx}_{i, \veps_i}), \tilde{\vx}_{i, \veps_i} \in \gA_{\veps_i,\vtheta})}_{\text{Type \RNum{1} adversarial examples on } \Ds} \\
    &+ \underbrace{\ind(\vx_i\in \Ds)\ind(\htt(\tilde{\vx}_{i, \veps_i})\neq\hbd(\tilde{\vx}_{i, \veps_i}), \tilde{\vx}_{i, \veps_i} \in \gA_{\veps_i,\vtheta})}_{\text{Type \RNum{2} or \RNum{3} adversarial examples on } \Ds} + \underbrace{\ind(\vx_i\in \Dn\setminus\Ds, \tilde{\vx}_{i, \veps_i} \in \gA_{\veps_i,\vtheta})}_{\text{vanilla  adversarial examples on } \Dn\setminus\Ds}\Big\}.
\end{aligned}   
\end{equation}

Proposition~\ref{ass:share} implies that the first component in $(\ref{eq::decom})$ is essential for backdoor mitigation, as it captures the risk of shared adversarial examples (SAEs) between the poisoned model $\hbd$ and the fine-tuned model $\htt$, and the poisoned samples effective for $\htt$ belong to SAEs. The second component is irrelevant for backdoor mitigation, since adversarial examples of Type \RNum{2} and \RNum{3}
are either not poisoned samples or ineffective to backdoor attack against $\htt$. The third component protects the samples that are originally resistant to triggers from being backdoor attacked in the mitigation process.

Thus, by removing the second component in $(\ref{eq::decom})$, we propose the following sub-adversarial risk
\begin{equation}
\label{eq::sha_risk}
\begin{aligned}
    \gR_{sub}(\htt) = &\frac{1}{|\Dn|} \sum_{i=1}^N \max_{\veps_i \in \gS} \Big\{\ind(\vx_i\in \Ds)\ind(\htt(\tilde{\vx}_{i, \veps_i})=\hbd(\tilde{\vx}_{i, \veps_i}), \tilde{\vx}_{i, \veps_i} \in \gA_{\veps_i,\vtheta}) \\
    & + \ind(\vx_i\in \Dn\setminus\Ds, \tilde{\vx}_{i, \veps_i} \in \gA_{\veps_i,\vtheta})\Big\}.
\end{aligned}   
\end{equation}

Compared to vanilla  adversarial risk $\gR_{adv}$, the proposed sub-adversarial risk $\gR_{sub}$ focuses on the shared adversarial examples for samples in $\Ds$ while considering vanilla  adversarial examples for samples in $\Dn\setminus\Ds$. When $\Ds=\Dn$, the sub-adversarial risk measures the shared adversarial risk on $\Dn$ and it is equivalent to $\gR_{adv}$ if $\Ds=\emptyset$.

Then, we provide the following proposition  to establish the relation between backdoor risk and (sub) adversarial risk:

\begin{proposition}    
Under Assumption \ref{assm:S}, for a classifier $\htt$, the following inequalities hold
$$\gR_{bd}(\htt) \leq  \gR_{sub}(\htt) \leq \gR_{adv}(\htt).$$
\end{proposition}

Therefore, $\gR_{sub}$ is a tighter upper bound for $\gR_{bd}$ compared with $\gR_{adv}$. 
After replacing $\gR_{bd}$ with $\gR_{sub}$ in Problem (\ref{eq::trade}), we reach the following optimization problem:
\begin{equation}
    \label{eq::sub_risk}
    \min_{\vtheta} \gR_{cl}(\htt) + \lambda \gR_{sub}(\htt).
\end{equation}
Due to the space limit, the proofs of above propositions, as well as some detailed discussions and extensions will be provided in \app. 

\subsection{Proposed method}
\label{sec::loss}
Since the target label $\hat{y}$ is unavailable and 0-1 loss is non-differentiable, we first discuss the relaxation of $\gR_{sub}$ and then replace 0-1 loss with suitable surrogate loss functions in this section.

\paragraph{Relaxation.}  
One limitation of $\gR_{sub}$ is the inaccessibility of $\Ds$. Since a poisoned model usually has a high attack success rate (ASR), the first relaxation is replacing $\Ds$ with $\Dn$, \ie,
\begin{equation*}
        \gR_{sub}(\htt) \approx \frac{1}{|\Dn|} \sum_{i=1}^N \max_{\veps_i \in \gS} \Big\{\ind(\vx_i\in \Dn)\ind(\htt(\tilde{\vx}_{i, \veps_i})=\hbd(\tilde{\vx}_{i, \veps_i}), \tilde{\vx}_{i, \veps_i} \in \gA_{\veps_i,\vtheta})\Big\}.
\end{equation*}
Since $\hat{y}$ is unavailable, we also replace $\Dn$ by $\Db$ in the sub-adversarial risk. As $g(\vx;\Delta)$ is not harmful for classification if the ground truth label of $\vx$ is $\hat{y}$, this relaxation has negligible negative influence for mitigating backdoor risk. 
After the above two relaxations, we reach the following shared adversarial risk on $\Db$: 
\begin{equation*}
  \gR_{share}(\htt) = \frac{1}{N}\sum_{i=1}^N \max_{\veps_i\in \gS} \Big\{\ind(\htt(\tilde{\vx}_{i, \veps_i})= \hbd(\tilde{\vx}_{i, \veps_i}), \tilde{\vx}_{i, \veps_i} \in \gA_{\veps_i,\vtheta})\Big\}.  
\end{equation*}
Due to the space limit, we postpone the detailed discussion of the above two relaxations (\eg, the relaxation gaps) in \app.

By replacing $\gR_{sub}$ with $\gR_{share}$ in Problem (\ref{eq::sub_risk}), we reach the following problem:
\begin{equation}
    \label{eq::final_risk}
    \min_{\vtheta} \gR_{cl}(\htt) + \lambda \gR_{share}(\htt).
\end{equation}
By solving ($\ref{eq::final_risk}$), we seek  $\htt$ that either classifies the adversarial examples correctly (\ie, $\htt(\tilde{\vx}_{\veps})=y$), or differently with $\hbd$ (\ie, $\htt(\tilde{\vx}_{\veps})\neq \hbd(\tilde{\vx}_{\veps})$), while maintaining a high accuracy on $\Db$.
\paragraph{Approximation.} We firstly need to approximate two indicators in Problem (\ref{eq::final_risk}), including: 1) $\ind(\htt(\vx)\neq y)$; 2) $\ind(\tilde{\vx}_{\veps} \in \gA_{\veps,\vtheta}, \htt(\tilde{\vx}_{\veps})= \hbd(\tilde{\vx}_{\veps}))$.
The former indicator is for the clean accuracy, so we use the widely used cross-entropy (CE) loss as the surrogate loss, \ie, $L_{cl}(\vx,y;\vtheta) = \text{CE}(\vp(\vx;\vtheta),y)$. 
In terms of the latter indicator that measures the shared adversarial risk, different surrogate losses could be employed for different usages:
\begin{itemize}[leftmargin=*]
    \item To generate shared adversarial examples, we first decompose the second indicator as below:
    \begin{equation}
        \label{eq:cap}
      \ind(\tilde{\vx}_{\veps} \in \gA_{\veps,\vtheta}\cap \gA_{\veps,\vtheta_{bd}})\ind( \htt(\tilde{\vx}_{\veps}) = \hbd(\tilde{\vx}_{\veps})), 
    \end{equation}
    which covers two components standing for fooling $\hbd$ and $\htt$ simultaneously, and keeping the same predicted label for $\hbd$ and $\htt$, respectively. 

    For the first component, we use CE on both $\htt$ and $\hbd$ as a surrogate loss, \ie, 
    \begin{equation}
    L_{adv}(\tilde{\vx}_{\veps},y;\vtheta) = \frac{1}{2}\left(\text{CE}(\vp(\tilde{\vx}_{\veps};\vtheta),y)+ \text{CE}(\vp(\tilde{\vx}_{\veps};\vtheta_{bd}),y)\right).
    \end{equation}
    For the second component, we measure the distance between $\vp(\tilde{\vx}_{i, \veps_i}; \vtheta)$ and $\vp(\tilde{\vx}_{\veps};\vtheta_{bd})$ by Jensen-Shannon  divergence \cite{fuglede2004jensen} and adopt the following surrogate loss:
        \begin{equation}
        L_{share}(\tilde{\vx}_{\veps},y;\vtheta) = - \text{JS}(\vp(\tilde{\vx}_{\veps}; \vtheta),\vp(\tilde{\vx}_{\veps}; \vtheta_{bd})).
        \end{equation}
    
    \item To unlearn the shared adversarial examples, we adopt the following surrogate loss 
    \begin{equation}
    L_{sar}(\tilde{\vx}_{\veps},y;\vtheta) = - \ind\left( \tilde{y}\neq y\right)\log\left(1-\vp_{\tilde{y}}(\tilde{\vx}_{\veps};\vtheta)\right),
    \end{equation}
    where $\tilde{y}=\hbd(\tilde{\vx}_{\veps})$. By reducing $L_{asr}$, the prediction of $\tilde{\vx}_{\veps}$ by $\htt$, \ie, $\htt(\tilde{\vx}_{\veps})$ is forced to differ from $\hbd(\tilde{\vx}_{\veps})$ if $\hbd(\tilde{\vx}_{\veps})\neq y$, therefore, reducing the shared adversarial risk (SAR).
\end{itemize}
\textbf{Overall objective.} By combining all the above surrogate losses with the linear weighted summation mode, we propose the following bi-level objective:
\begin{equation}
\label{eq::overall}
\begin{aligned}
    \min_{\vtheta} \quad &\frac{1}{N} \sum_{i=1}^N \left\{  \lambda_1 L_{cl}(\vx_i,y_i;\vtheta) + \lambda_2  L_{sar}(\tilde{\vx}_{i,\veps_i^*},y_i;\vtheta)\right\}\\
    \mathrm{s.t.} \quad & \veps_i^* = \argmax_{\veps_i \in \gS} \lambda_3 L_{adv}(\tilde{\vx}_{i,\veps_i},y_i;\vtheta) +\lambda_4  L_{share}(\tilde{\vx}_{i,\veps_i},y_i;\vtheta), \quad i=1, \dots, N,  
\end{aligned}
\end{equation}
where $\lambda_1, \lambda_2, \lambda_3, \lambda_4 \geq 0$ indicate trade-off weights. 

\paragraph{Optimization algorithm.} We propose an optimization algorithm called \textbf{Shared Adversarial Unlearning} (SAU), which solves Problem ~(\ref{eq::overall}) by alternatively updating $\vtheta$, and $\veps^*$. Specifically, the model parameter $\vtheta$  is updated using stochastic  gradient descent \cite{SGD}, and the perturbation $\veps^*$ is generated by projected gradient descent \cite{madrytowards}. 
More details of SAU will be presented in \app.

\section{Experiment}
\subsection{Experiment settings}
\paragraph{Backdoor attack.} We compare SAU with 7 popular state-of-the-art (SOTA)  backdoor attacks, including BadNets \cite{gu2019badnets}, Blended backdoor attack (Blended) \cite{chen2017targeted}, input-aware dynamic backdoor attack (Input-Aware)\cite{nguyen2020input}, low frequency attack (LF) \cite{zeng2021rethinking}, sinusoidal signal backdoor attack (SIG) \cite{barni2019new}, sample-specific backdoor attack (SSBA) \cite{li2021invisible}, and warping-based poisoned networks (WaNet) \cite{nguyen2021wanet}. To make a fair and trustworthy comparison, we use the implementation and configuration from BackdoorBench \cite{wubackdoorbench}, a comprehensive benchmark for backdoor evaluation, such as the trigger patterns and optimization hyper-parameters. By default, the poisoning ratio is set to 10\% in all attacks, and the $0^{th}$ label is set to be the target label. We evaluate all the attacks on 3 benchmark datasets, CIFAR-10 \cite{krizhevsky2009learning}, Tiny ImageNet \cite{le2015tiny}, and GTSRB \cite{stallkamp2011german}  using two networks: PreAct-ResNet18 \cite{he2016identity} and VGG19 \cite{simonyan2014very}. Due to space constraints, the results for GTSRB and VGG19 are postponed to \app. Note that for clean label attack SIG, the 10\% poisoning ratio can only be implemented for CIFAR-10. More attack details are left in \app.

\paragraph{Backdoor defense.}
We compare our method with 6 SOTA backdoor defense methods: ANP \cite{wu2021adversarial}, Fine-pruning (FP) \cite{liu2018fine}, NAD \cite{li2021neural}, NC \cite{wang2019neural}, EP \cite{zheng2022preactivation} and i-BAU \cite{zeng2022adversarial}. By default, all the defense methods can access 5\% benign training data. We follow the recommended configurations for SOTA defenses as in BackdoorBench \cite{wubackdoorbench}. For our method, we choose to generate the shared Adversarial example with Projected Gradient Descent (PGD) \cite{madrytowards} with $L_{\infty}$ norm. For all experiments, we run PGD 5 steps with norm bound $0.2$ 
and we set $\lambda_1=\lambda_2=\lambda_4=1$ and $\lambda_3 = 0.01$. More details about defense settings and additional experiments can be found in  \app.

\paragraph{Evaluation metric.}  We use four metrics to evaluate the performance of each defense method: Accuracy on benign data ($\textbf{ACC}$), Attack Success Rate ($\textbf{ASR}$), Robust Accuracy ($\textbf{R-ACC}$) and Defense Effectiveness Rating ($\textbf{DER}$). R-ACC measures the proportion of the poisoned samples classified to their true label, and ASR measures the proportion of poisoned samples misclassified to the target label. Larger R-ACC and lower ASR indicate that the backdoor is effectively mitigated. Note that the samples for the target class are excluded when computing the ASR and R-ACC as done in BackdoorBench. DER $\in [0,1] $ was proposed in \cite{zhu2023enhancing, zhu2023neural} to evaluate the cost of ACC for reducing ASR. It is defined as follows:
\begin{equation}
    \text{DER} = [\max(0, \Delta \text{ASR})-\max(0,\Delta \text{ACC})+1]/2,
\end{equation}
where $\Delta \text{ASR}$ denotes the drop in ASR after applying defense, and $\Delta \text{ACC}$ represents the drop in ACC after applying defense.

\textbf{Note}: Higher ACC, lower ASR, higher R-ACC and higher DER represent better defense performance. We use \textbf{boldface} and \underline{underline}  to indicate the best and the second-best results among all defense methods, respectively, in later experimental results.

\subsection{Main results}

\paragraph{Effectiveness of SAU.} To verify the effectiveness of  SAU, we first summarize the experimental results on CIFAR-10 and Tiny ImageNet in Tables \ref{cifar10_preactresnet18_1} and \ref{tiny_preactresnet18_1}, respectively. As shown in Tables \ref{cifar10_preactresnet18_1}-\ref{tiny_preactresnet18_1},  SAU can mitigate backdoor for almost all attacks with a significantly lower average ASR. For the experiments on CIFAR-10,  SAU achieves the top-2 lowest ASR in 4 of 7 attacks and very low ASR for the other 3 attacks. Similarly,  SAU performs the lowest ASR in 3 attacks for Tiny ImageNet and negligible ASR in two of the other attacks. Notably,  SAU fails to mitigate WaNet in Tiny ImageNet, although it can defend against WaNet on CIFAR-10. As a transformation-based attack that applies image transforms to construct poisoned samples, WaNet can generate triggers with a large $L_{\infty}$ norm.  Specifically,  WaNet has average $L_\infty$ trigger norm $0.348$ on TinyImageNet and $0.172$ on CIFAR-10. Note that for all experiments, we generate adversarial perturbation with $L_\infty$ norm less than $0.2$. The large trigger norm of WaNet on Tiny ImageNet poses a challenge for AT-based methods such as I-BAU and the proposed one, which reveals an important weakness of such methods, \ie, their performance may be degraded if the trigger is beyond the perturbation set. 

As maintaining clean accuracy is also important for an effective backdoor defense method, we also compare  SAU with other baselines in terms of ACC, R-ACC and DER in Table~\ref{cifar10_preactresnet18_1} and \ref{tiny_preactresnet18_1}. As  SAU adopts adversarial examples to mitigate backdoors, it has negative effects on clean accuracy, resulting in a slightly lower clean accuracy compared to the best one. However, SAU achieves the best average R-ACC, which demonstrates its effectiveness in recovering the prediction of poisoned samples, \ie, classifying the poisoned samples correctly. Besides, the best average DER  indicates that  SAU achieves a significantly better tradeoff between clean accuracy and backdoor risk.

\begin{table}[t]
\centering
\caption{Results on CIFAR-10 with PreAct-ResNet18 and poisoning ratio $10\%$.}
\label{cifar10_preactresnet18_1}
\setlength{\tabcolsep}{3pt} 
\scalebox{0.68}{
\begin{tabular}{c|cccc|cccc|cccc|cccc}
\toprule
Defense & \multicolumn{4}{c|}{No Defense}    & \multicolumn{4}{c|}{ANP \cite{wu2021adversarial}}    & \multicolumn{4}{c|}{FP \cite{liu2018fine}}& \multicolumn{4}{c}{NC \cite{wang2019neural}}   \\ \midrule
Attack  & \multicolumn{1}{c}{ACC} & \multicolumn{1}{c}{ASR} & \multicolumn{1}{c}{R-ACC} & \multicolumn{1}{c|}{DER}& \multicolumn{1}{c}{ACC}  & \multicolumn{1}{c}{ASR} & \multicolumn{1}{c}{R-ACC} & \multicolumn{1}{c|}{DER}  & \multicolumn{1}{c}{ACC}         & \multicolumn{1}{c}{ASR}  & \multicolumn{1}{c}{R-ACC}& \multicolumn{1}{c|}{DER}  & \multicolumn{1}{c}{ACC}  & \multicolumn{1}{c}{ASR} & \multicolumn{1}{c}{R-ACC}& \multicolumn{1}{c}{DER}  \\
\midrule
BadNets \cite{gu2019badnets} & $91.32$& $95.03$& $4.67$&  N/A & $\underline{90.88}$& $4.88$& $87.22$& $94.86$& $\textbf{91.31}$& $57.13$& $41.62$& $68.95$& $89.05$& $\underline{1.27}$& $\underline{89.16}$& $95.75$\\
Blended \cite{chen2017targeted} & $93.47$& $99.92$& $0.08$&  N/A & $92.97$& $84.88$& $13.36$& $57.27$& $\underline{93.17}$& $99.26$& $0.73$& $50.18$& $\textbf{93.47}$& $99.92$& $0.08$& $50.00$\\
Input-Aware \cite{nguyen2020input} & $90.67$& $98.26$& $1.66$&  N/A & $91.04$& $1.32$& $86.71$& $98.47$& $91.74$& $\textbf{0.04}$& $44.54$& $\textbf{99.11}$& $\underline{92.61}$& $\underline{0.76}$& $\underline{90.87}$& $\underline{98.75}$\\
LF \cite{zeng2021rethinking} & $93.19$& $99.28$& $0.71$&  N/A & $\underline{92.64}$& $39.99$& $55.03$& $79.37$& $\textbf{92.90}$& $98.97$& $1.02$& $50.01$& $91.62$& $\textbf{1.41}$& $\textbf{87.48}$& $\textbf{98.15}$\\
SIG \cite{barni2019new} & $84.48$& $98.27$& $1.72$&  N/A & $83.36$& $36.42$& $43.67$& $80.36$& $\underline{89.10}$& $26.20$& $20.61$& $86.03$& $84.48$& $98.27$& $1.72$& $50.00$\\
SSBA \cite{li2021invisible} & $92.88$& $97.86$& $1.99$&  N/A & $\textbf{92.62}$& $60.17$& $36.69$& $68.71$& $\underline{92.54}$& $83.50$& $15.36$& $57.01$& $90.99$& $\textbf{0.58}$& $\textbf{87.04}$& $\textbf{97.69}$\\
WaNet \cite{nguyen2021wanet} & $91.25$& $89.73$& $9.76$&  N/A & $91.33$& $2.22$& $88.54$& $93.76$& $91.46$& $\underline{1.09}$& $69.73$& $\underline{94.32}$& $\underline{91.80}$& $7.53$& $85.09$& $91.10$\\
Average & $91.04$& $96.91$& $2.94$&  N/A & $90.69$& $32.84$& $58.75$& $81.83$& $\underline{91.75}$& $52.31$& $27.66$& $72.23$& $90.57$& $29.96$& $63.06$& $83.06$\\

\toprule

\toprule
Defense & \multicolumn{4}{c|}{NAD \cite{li2021neural}}  & \multicolumn{4}{c|}{EP \cite{zheng2022preactivation} } & \multicolumn{4}{c|}{i-BAU \cite{zeng2022adversarial}}        & \multicolumn{4}{c}{ SAU (\textbf{Ours})} \\
\midrule
Attack  & \multicolumn{1}{c}{ACC} & \multicolumn{1}{c}{ASR} & \multicolumn{1}{c}{R-ACC} & \multicolumn{1}{c|}{DER}  &  \multicolumn{1}{c}{ACC} & \multicolumn{1}{c}{ASR} & \multicolumn{1}{c}{R-ACC} & \multicolumn{1}{c|}{DER}  & \multicolumn{1}{c}{ACC}         & \multicolumn{1}{c}{ASR}  & \multicolumn{1}{c}{R-ACC}& \multicolumn{1}{c|}{DER}  & \multicolumn{1}{c}{ACC}  & \multicolumn{1}{c}{ASR} & \multicolumn{1}{c}{R-ACC}& \multicolumn{1}{c}{DER}  \\
\midrule
BadNets \cite{gu2019badnets} & $89.87$& $2.14$& $88.71$& $95.72$& $89.66$& $1.88$& $\textbf{89.51}$& $\underline{95.75}$& $89.15$& $\textbf{1.21}$& $88.88$& $\textbf{95.83}$& $89.31$& $1.53$& $88.81$& $95.74$\\
Blended \cite{chen2017targeted} & $92.17$& $97.69$& $2.14$& $50.47$& $92.43$& $52.13$& $37.52$& $73.37$& $88.66$& $\underline{13.99}$& $\underline{53.23}$& $\underline{90.56}$& $90.96$& $\textbf{6.14}$& $\textbf{64.89}$& $\textbf{95.63}$\\
Input-Aware \cite{nguyen2020input} & $\textbf{93.18}$& $1.68$& $\textbf{91.12}$& $98.29$& $89.86$& $2.23$& $85.20$& $97.61$& $90.29$& $63.36$& $32.70$& $67.26$& $91.59$& $1.27$& $88.54$& $98.49$\\
LF \cite{zeng2021rethinking} & $92.37$& $47.83$& $47.49$& $75.31$& $91.82$& $85.98$& $12.77$& $55.97$& $89.09$& $21.83$& $64.37$& $86.67$& $90.32$& $\underline{4.18}$& $\underline{81.54}$& $\underline{96.12}$\\
SIG \cite{barni2019new} & $\textbf{90.02}$& $10.66$& $\textbf{64.20}$& $93.81$& $83.1$& $\textbf{0.26}$& $56.68$& $\underline{98.32}$& $85.85$& $\underline{1.28}$& $55.19$& $\textbf{98.49}$& $88.56$& $1.67$& $\underline{57.96}$& $98.30$\\
SSBA \cite{li2021invisible} & $91.91$& $77.4$& $20.86$& $59.74$& $92.33$& $10.67$& $78.60$& $93.32$& $88.15$& $2.17$& $77.28$& $95.48$& $90.84$& $\underline{1.79}$& $\underline{85.83}$& $\underline{97.01}$\\
WaNet \cite{nguyen2021wanet} & $\textbf{93.17}$& $22.98$& $72.69$& $83.38$& $90.09$& $86.64$& $12.54$& $50.96$& $90.91$& $3.37$& $\underline{89.10}$& $93.01$& $91.26$& $\textbf{1.02}$& $\textbf{90.28}$& $\textbf{94.36}$\\
Average & $\textbf{91.81}$& $37.2$& $55.32$& $79.53$& $89.9$& $34.26$& $53.26$& $80.76$& $88.87$& $\underline{15.32}$& $\underline{65.82}$& $\underline{89.61}$& $90.41$& $\textbf{2.51}$& $\textbf{79.69}$& $\textbf{96.52}$\\

\bottomrule
\end{tabular}}
\end{table}

\begin{table}[t]
\centering
\caption{Results on Tiny ImageNet with PreAct-ResNet18 and poisoning ratio $10\%$.}
\label{tiny_preactresnet18_1}
\setlength{\tabcolsep}{3pt} 
\scalebox{0.68}{
\begin{tabular}{c|cccc|cccc|cccc|cccc}
\toprule
Defense & \multicolumn{4}{c|}{No Defense}    & \multicolumn{4}{c|}{ANP \cite{wu2021adversarial}}    & \multicolumn{4}{c|}{FP \cite{liu2018fine}}& \multicolumn{4}{c}{NC \cite{wang2019neural}}   \\ \midrule
Attack  & \multicolumn{1}{c}{ACC} & \multicolumn{1}{c}{ASR} & \multicolumn{1}{c}{R-ACC} & \multicolumn{1}{c|}{DER}& \multicolumn{1}{c}{ACC}  & \multicolumn{1}{c}{ASR} & \multicolumn{1}{c}{R-ACC} & \multicolumn{1}{c|}{DER}  & \multicolumn{1}{c}{ACC}         & \multicolumn{1}{c}{ASR}  & \multicolumn{1}{c}{R-ACC}& \multicolumn{1}{c|}{DER}  & \multicolumn{1}{c}{ACC}  & \multicolumn{1}{c}{ASR} & \multicolumn{1}{c}{R-ACC}& \multicolumn{1}{c}{DER}  \\
\midrule
BadNets \cite{gu2019badnets} & $56.23$& $100.0$& $0.0$&  N/A & $43.45$& $\textbf{0.00}$& $43.13$& $93.61$& $51.73$& $99.99$& $0.01$& $47.76$& $51.52$& $0.10$& $50.82$& $\underline{97.59}$\\
Blended \cite{chen2017targeted} & $56.03$& $99.71$& $0.22$&  N/A & $43.93$& $\underline{6.11}$& $\underline{17.28}$& $\underline{90.75}$& $51.89$& $95.94$& $2.1$& $49.81$& $\underline{52.55}$& $93.21$& $3.96$& $51.51$\\
Input-Aware \cite{nguyen2020input} & $57.45$& $98.85$& $1.06$&  N/A & $35.39$& $\textbf{0.00}$& $11.82$& $88.40$& $55.28$& $62.92$& $24.7$& $66.88$& $\underline{56.20}$& $\underline{0.09}$& $\textbf{52.19}$& $\underline{98.76}$\\
LF \cite{zeng2021rethinking} & $55.97$& $98.57$& $0.97$&  N/A & $45.69$& $62.30$& $14.49$& $63.00$& $51.44$& $95.25$& $2.42$& $49.4$& $\underline{52.99}$& $85.56$& $8.07$& $55.02$\\
SSBA \cite{li2021invisible} & $55.22$& $97.71$& $1.68$&  N/A & $43.36$& $56.53$& $17.24$& $64.66$& $50.47$& $88.87$& $6.26$& $52.04$& $\underline{52.47}$& $53.47$& $\underline{23.17}$& $70.75$\\
WaNet \cite{nguyen2021wanet} & $56.78$& $99.49$& $0.36$&  N/A & $36.16$& $\textbf{0.07}$& $31.97$& $89.40$& $53.84$& $3.94$& $2.38$& $96.3$& $53.33$& $0.23$& $\underline{51.63}$& $\underline{97.90}$\\
Average & $56.28$& $99.06$& $0.72$&  N/A & $41.33$& $\underline{20.83}$& $22.66$& $77.12$& $52.44$& $74.48$& $6.31$& $58.88$& $\underline{\textbf{53.18}}$& $38.78$& $\underline{31.64}$& $74.50$\\

\toprule

\toprule
Defense & \multicolumn{4}{c|}{NAD \cite{li2021neural}}  & \multicolumn{4}{c|}{EP \cite{zheng2022preactivation}} & \multicolumn{4}{c|}{i-BAU \cite{zeng2022adversarial}}        & \multicolumn{4}{c}{ SAU (\textbf{Ours})} \\
\midrule
Attack  & \multicolumn{1}{c}{ACC} & \multicolumn{1}{c}{ASR} & \multicolumn{1}{c}{R-ACC} & \multicolumn{1}{c|}{DER}  &  \multicolumn{1}{c}{ACC} & \multicolumn{1}{c}{ASR} & \multicolumn{1}{c}{R-ACC} & \multicolumn{1}{c|}{DER}  & \multicolumn{1}{c}{ACC}         & \multicolumn{1}{c}{ASR}  & \multicolumn{1}{c}{R-ACC}& \multicolumn{1}{c|}{DER}  & \multicolumn{1}{c}{ACC}  & \multicolumn{1}{c}{ASR} & \multicolumn{1}{c}{R-ACC}& \multicolumn{1}{c}{DER}  \\
\midrule
BadNets \cite{gu2019badnets} & $46.37$& $0.27$& $45.61$& $94.93$& $\underline{52.30}$& $\underline{0.03}$& $\textbf{52.05}$& $\textbf{98.02}$& $\textbf{52.67}$& $98.31$& $1.59$& $49.06$& $51.52$& $0.53$& $\underline{51.15}$& $97.38$\\
Blended \cite{chen2017targeted} & $46.89$& $94.99$& $2.63$& $47.79$& $51.86$& $56.03$& $12.31$& $69.75$& $\textbf{52.73}$& $92.96$& $4.15$& $51.72$& $50.30$& $\textbf{0.06}$& $\textbf{25.27}$& $\textbf{96.96}$\\
Input-Aware \cite{nguyen2020input} & $47.91$& $1.86$& $\underline{43.13}$& $93.73$& $\textbf{57.23}$& $0.24$& $24.74$& $\textbf{99.20}$& $50.63$& $56.12$& $24.86$& $67.96$& $54.11$& $0.33$& $42.29$& $97.59$\\
LF \cite{zeng2021rethinking} & $45.45$& $\underline{50.49}$& $\underline{20.21}$& $\underline{68.78}$& $\textbf{53.33}$& $75.41$& $13.06$& $60.26$& $52.77$& $88.04$& $6.82$& $53.67$& $52.65$& $\textbf{0.97}$& $\textbf{35.87}$& $\textbf{97.14}$\\
SSBA \cite{li2021invisible} & $45.32$& $57.32$& $19.44$& $65.25$& $48.13$& $\underline{44.07}$& $21.98$& $\underline{73.27}$& $\textbf{52.52}$& $74.77$& $14.31$& $60.12$& $51.85$& $\textbf{0.11}$& $\textbf{36.36}$& $\textbf{97.11}$\\
WaNet \cite{nguyen2021wanet} & $46.98$& $0.43$& $43.99$& $94.63$& $\textbf{56.21}$& $\underline{0.23}$& $\textbf{55.13}$& $\textbf{99.34}$& $54.40$& $94.86$& $3.92$& $51.12$& $\underline{54.65}$& $85.75$& $10.35$& $55.80$\\
Average & $46.49$& $34.23$& $29.17$& $73.59$& $53.18$& $29.33$& $29.88$& $\underline{78.55}$& $52.62$& $84.18$& $9.28$& $54.81$& $52.51$& $\textbf{14.62}$& $\textbf{33.55}$& $\textbf{84.57}$\\
\bottomrule
\end{tabular}}
\end{table}

\begin{table}[h]
\centering
\caption{Results on CIFAR-10 with PreAct-ResNet18 and different poisoning ratios.}
\label{tab:pratio}
\setlength{\tabcolsep}{3pt} 
\scalebox{0.68}{
\begin{tabular}{c|c|cc|cc|cc|cc|cc}
\toprule
\multicolumn{2}{c|}{Poisoning Ratio} & \multicolumn{2}{c|}{10\%} & \multicolumn{2}{c|}{20\%} & \multicolumn{2}{c|}{30\%} & \multicolumn{2}{c|}{40\%} & \multicolumn{2}{c}{50\%} \\ \midrule
Attack                   & Metric & No Defence    & Ours      & No Defence    & Ours      & No Defence    & Ours      & No Defence    & Ours      & No Defence    & Ours      \\
\midrule
\multirow{2}{*}{BadNets \cite{gu2019badnets}}  & ACC    & 91.82         & 89.05    & 90.17         & 89.27    & 88.32         & 88.49    & 86.16         & 87.30    & 83.61         & 86.93    \\
& ASR    & 93.79         & 1.53     & 96.12         & 1.17     & 97.33         & 1.72     & 97.78         & 1.72     & 98.39         & 2.31     \\
\midrule
\multirow{2}{*}{Blended \cite{chen2017targeted}} & ACC    & 93.69         & 90.96    & 93.00         & 91.02    & 92.78         & 89.49    & 91.64         & 89.64    & 91.26         & 88.92    \\
& ASR    & 99.76         & 6.14     & 99.92         & 7.99     & 99.98         & 11.58    & 99.96         & 5.36     & 100.00        & 4.63     \\
\midrule
\multirow{2}{*}{LF \cite{zeng2021rethinking}}      & ACC    & 93.01         & 90.32    & 92.60         & 91.33    & 92.01         & 89.65    & 92.03         & 89.09    & 90.91         & 89.02    \\
& ASR    & 99.06         & 4.18     & 99.54         & 1.90     & 99.76         & 2.02     & 99.84         & 1.94     & 99.89         & 1.77     \\
\bottomrule
\end{tabular}}
\end{table}

\paragraph{Influence of poisoning ratio.}
To study  the influence of the poisoning ratio on the effectiveness of  SAU, we test  SAU with varying poisoning ratios from 10\% to 50\%. As shown in Table \ref{tab:pratio},  SAU is still able to achieve remarkable performance even if half of the data is poisoned.

\subsection{Ablation Study}
\label{sec:abl}
\begin{figure}[h]
    \centering
    \includegraphics[width=\linewidth]{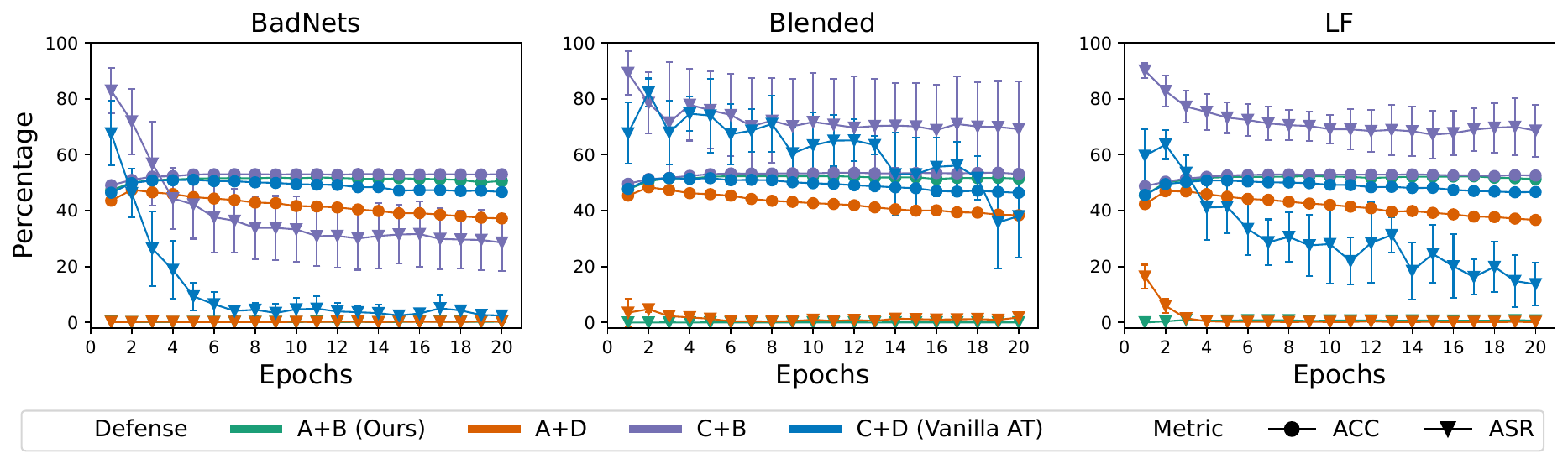}
    \caption{Learn curves for SAU, Vanilla AT and their variants, averaged over five runs.}
    \label{fig:abl}
\end{figure}

In this section, we conduct ablation study to investigate each component of  SAU. Specifically, SAU is composed of two parts: generating shared adversarial examples according to the maximization step in (\ref{eq::overall}) denoted by \textbf{A}, and reducing shared adversarial risk according to the minimization step in  (\ref{eq::overall}), denote by \textbf{B}. To study the effect of these parts, we replace them with the corresponding part in vanilla adversarial training, denoted by \textbf{C} and \textbf{D}, respectively. Then, we run each variant $20$ epochs on Tiny ImageNet with PreAct-ResNet18 and a poisoning ratio $10\%$. Each experiment is repeated five times, and the error bar is only shown in Figure~\ref{fig:abl} for simplicity.

\begin{wraptable}{R}{6.5cm}
\caption{Results on Tiny ImageNet with different variants of SAU and Vanilla AT.}   	\label{abl}
\setlength{\tabcolsep}{3pt} 
\scalebox{0.72}{
\begin{tabular}{c|cc|cc|cc}
\toprule
Attack     & \multicolumn{2}{c|}{BadNets \cite{gu2019badnets}} & \multicolumn{2}{c|}{Blended \cite{chen2017targeted}} & \multicolumn{2}{c}{LF \cite{zeng2021rethinking}} \\ \midrule
Defense    & ACC           & ASR          & ACC           & ASR          & ACC        & ASR       \\ \midrule
\textbf{A}+\textbf{B} (Ours) &  $ 50.60 $ & $ 0.27 $ &       $ 51.47 $ & $ 0.02 $ &        $ 51.31 $ & $ 0.72 $     \\ \midrule
\textbf{A}+\textbf{D}        &  $ 37.21 $ & $ 0.30 $ &        $ 38.35 $ & $ 1.77 $ &       $ 36.68 $ & $ 0.17 $      \\ \midrule
\textbf{C}+\textbf{B}        & $ 53.05 $ & $ 28.57 $ &        $ 53.16 $ & $ 69.22 $ &       $ 52.57 $ & $ 68.65 $     \\ \midrule
\textbf{C}+\textbf{D} (Vanilla AT)   & $ 46.66 $ & $ 2.48 $ &        $ 46.39 $ & $ 38.03 $ &  $ 46.69 $ & $ 13.68 $    \\ \bottomrule
\end{tabular}}
\end{wraptable}
 
As shown in Table~\ref{abl}, the maximization step for generating shared adversarial examples (component \textbf{A}) is the key to reducing ASR, and component \textbf{B} can help to alleviate the hurt to clean accuracy. When replacing the component \textbf{B} with the minimization step in vanilla AT (\textbf{D}), it suffers from a drop of clean accuracy, although it can also work for mitigating backdoor effectively. Compared to vanilla AT,  SAU achieves significantly lower ASR and higher ACC, with a much more stable learning curve, as shown in Figure~\ref{fig:abl}. We remark that the decrease of clean accuracy for \textbf{A}+\textbf{D} and vanilla AT   demonstrates that imposing a model to learn overwhelming adversarial examples may hurt the clean accuracy, a phenomenon widely studied  in the area of adversarial training \cite{tsiprasrobustness, zhang2019theoretically}.

\section{Conclusion}

In conclusion, this paper proposes Shared Adversarial Unlearning, a method to defend against backdoor attacks in deep neural networks through adversarial training techniques. By developing a better understanding of the connection between adversarial examples and poisoned samples, we propose a novel upper bound for backdoor risk and a bi-level formulation for mitigating backdoor attacks in poisoned models. Our approach identifies shared adversarial examples and unlearns them to break the connection between the poisoned sample and the target label. We demonstrate the effectiveness of our proposed method through extensive experiments, showing that it achieves comparable to, and even superior, performance than six different state-of-the-art defense methods on seven SOTA backdoor attacks with different model structures and datasets. Our work provides a valuable contribution to the field of backdoor defense in deep neural networks, with potential applications in various domains.

\paragraph{Limitation and future work.} One important direction for future work, and a current challenge, is to defend against attacks with a large norm. A valuable approach for this challenge is to consider adversarial examples for the union of multiple perturbation types \cite{maini2022perturbation, tramer2019adversarial}. Another promising direction is to exploit the properties of backdoor, \eg, scaled prediction consistency \cite{guoscale} and latent separability \cite{qi2023revisiting},  to further study the connection between adversarial examples and poisoned samples. 
\newpage
\bibliographystyle{plainnat}
\bibliography{references}
\clearpage
\clearpage

\appendix

\section{Proofs and extension}

\subsection{Proof of Proposition 1}
\begin{proof}
Under Assumption 1, for any sample $\vx \in \Dn$, we have 
\begin{equation*}
    \ind(\htt(g(\vx;\Delta))=\hat{y}) \leq \ind(g(\vx;\Delta)\in\gA_{g(\vx;\Delta)-\vx, \vtheta}) \leq \max_{\veps\in\gS} \ind(\tilde{\vx}_{\veps}\in\gA_{\veps, \vtheta}),
\end{equation*}
since $g(\vx;\Delta) - \vx \in \gS$. 

Therefore, we have 
\begin{equation*}
  \gR_{bd}(\htt) \leq  \gR_{adv}(\htt).  
\end{equation*}
\end{proof}

\subsection{Proof of Proposition 2}
\begin{proof}
By the definition of backdoor attack, a poisoned sample $g(\vx;\Delta)$ is said to attack $\htt$ successfully if and only if $\htt(g(\vx;\Delta))=\hat{y}$. 

As $\hbd(g(\vx;\Delta))=\hat{y}$ for all $\vx\in \Ds$ (definition of $\Ds$), we have
$\htt(g(\vx;\Delta)) = \hbd(g(\vx;\Delta))\neq y$ if and only if $g(\vx;\Delta)$ can attack $\htt$ successfully, for all $\vx\in \Ds$.

Furthermore, under Assumption 1, $g(\vx;\Delta)$ is an shared adversarial example of $\htt$ and $\hbd$ if  $\htt(g(\vx;\Delta)) = \hbd(g(\vx;\Delta))\neq y$ by the Definition 1.

\end{proof}

\subsection{Proof of Proposition 3}
\begin{proof}
By Proposition 1 and the decomposition of $\gR_{adv}$ in Section 3.2, we have
\begin{align*}
  \gR_{bd}(\htt) &\leq \gR_{adv}(\htt),\\  
  \gR_{sub}(\htt) &\leq \gR_{adv}(\htt).
\end{align*}
Therefore, we discuss the relation between $\gR_{bd}(\htt) $ and $\gR_{sub}(\htt)$ now.

Recall that 
\begin{equation}
\begin{aligned}
    \gR_{sub}(\htt) = &\frac{1}{|\Dn|} \sum_{i=1}^N \max_{\veps_i \in \gS} \Big\{\ind(\vx_i\in \Ds)\ind(\htt(\tilde{\vx}_{i, \veps_i})=\hbd(\tilde{\vx}_{i, \veps_i}), \tilde{\vx}_{i, \veps_i} \in \gA_{\veps_i,\vtheta}) \\
    & + \ind(\vx_i\in \Dn\setminus\Ds, \tilde{\vx}_{i, \veps_i} \in \gA_{\veps_i,\vtheta})\Big\}.
\end{aligned}   
\end{equation}

Then, under Assumption 1, we have
\begin{itemize}[leftmargin=*]
    \item For $\vx \in \Ds$, Proposition 2 yields that 
    \begin{equation}
    \begin{aligned}
        \ind(\htt(g(\vx;\Delta))=\hat{y}) &= \ind(\htt(g(\vx;\Delta))=\hbd(g(\vx;\Delta))) \\ &\leq  \max_{\veps\in\gS} \ind(\htt(\tilde{\vx}_{\veps})=\hbd(\tilde{\vx}_{\veps}), \tilde{\vx}_{\veps}\in\gA_{\veps, \vtheta}).   
    \end{aligned}
    \end{equation}
    \item For $\vx\in \Dn\setminus\Ds$, similar as Proposition 1, we have 
    \begin{equation*}
            \ind(\htt(g(\vx;\Delta))=\hat{y}) \leq \ind(g(\vx;\Delta)\in\gA_{g(\vx;\Delta)-\vx, \vtheta}) \leq \max_{\veps\in\gS} \ind(\tilde{\vx}_{\veps}\in\gA_{\veps, \vtheta}).
    \end{equation*}
\end{itemize}

Therefore, we have
$$\gR_{bd}(\htt)  \leq \gR_{sub}(\htt).$$

So, under Assumption 1, the following inequalities holds:
\begin{align*}
  \gR_{bd}(\htt) \leq   \gR_{sub}(\htt) \leq \gR_{adv}(\htt).
\end{align*}
\end{proof}

\subsection{Extension to Universal Adversarial Perturbation}
Universal Adversarial Perturbation (UAP) has achieved remarkable performance in previous studies \cite{zeng2022adversarial, chai2022oneshot}, especially for defending against backdoor attacks with fixed additive triggers. Now, we extend $\gR_{sub}$ to UAP and show that shared adversarial examples also benefit adversarial training with UAP.

We denote the Universal Adversarial Risk by 
\begin{align*}
    \gR_{uadv}(\htt) = \frac{\max_{\veps \in \gS} \sum_{i=1}^N \ind(\tilde{\vx}_{i, \veps} \in \gA_{\veps,\vtheta}, \vx_i\in\Dn)}{|\Dn|}.
\end{align*}

To bridge UAP and backdoor trigger, we first provide the following assumption:
\begin{appass}
\label{ass:add}
Assume that the $g(\vx,\Delta) = \vx+\Delta $ for all $\vx\in \Db$.
\end{appass}

Assumption~\ref{ass:add} says that the poisoned samples are planted with a fixed additive trigger. 

Then, under Assumption~\ref{ass:add} and Assumption 1, we have
\begin{equation*}
    \gR_{bd}(\htt)\leq \gR_{uadv}\leq \gR_{adv}.
\end{equation*}
by the fact that $\Delta \in \gS$.

For clarity, we replace $\tilde{\vx}_{\veps}\in \gA_{\veps,\vtheta}$ by $\htt(\vx+\veps)\neq y$ and $\tilde{\vx}_{\veps}$ by $\vx+\veps$ which makes the rest derivation easier to follow. 

Then, similar as $\gR_{sub}$, we define the universal sub-adversarial risk $\gR_{usub}$ as below:
\begin{align*}
  \gR_{usub}(\htt) = \frac{1}{|\Dn|}\max_{\veps\in \gS} \sum_{i=1}^N  &\big\{\ind(\htt(\vx+\veps) = \hbd(\vx+\veps)\neq y,\vx\in \Ds)+\\
  &\ind(\htt(\vx+\veps)\neq y,\vx\in \Dn\setminus\Ds)\big\},  
\end{align*}

which considers  shared universal adversarial risk on $\Ds$ and vanilla universal adversarial risk on $\Dn\setminus\Ds$.

By the definition of $\gR_{usub}$, we can easily find $\gR_{usub}\leq \gR_{uadv}$.

Furthermore, we have

\resizebox{\linewidth}{!}{
\begin{minipage}{\linewidth}
\begin{align*}
        \gR_{bd}(\htt) &= \frac{\sum_{i=1}^N \ind(\htt(\vx+\Delta)=\hat{y},\vx\in \Dn)}{|\Dn|} \\
        & = \frac{\sum_{i=1}^N \left\{\ind(\htt(\vx+\Delta)=\hbd(\vx+\Delta),\vx\in \Ds)+\ind(\htt(\vx+\Delta)=\hat{y},\vx\in \Dn\setminus\Ds)\right\}}{|\Dn|}\\
        & \leq \frac{\max_{\veps\in \gS}\sum_{i=1}^N \left\{\ind(\htt(\vx+\veps) = \hbd(\vx+\veps),\vx\in \Ds)+\ind(\htt(\vx+\veps)\neq y,\vx\in \Dn\setminus\Ds)\right\}}{|\Dn|}\\
        & = \gR_{usub}(\htt).
\end{align*}
\end{minipage}}

Therefore, we reach the following proposition
\begin{appprop}
Under Assumption~\ref{ass:add} and Assumption 1, for a classifier $\htt$, the following inequalities hold
$$\gR_{bd}(\htt)\leq \gR_{usub}(\htt)\leq \gR_{uadv}(\htt)\leq \gR_{adv}(\htt).$$
\end{appprop}
Therefore, the shared adversarial examples also benefit adversarial training with UAP.

\subsection{Extension to Targeted Adversarial Training}
In the previous sections, we assume that the defender has no access to either the trigger or the target label. Since there are already some works for backdoor trigger detection \cite{guo2022aeva, dong2021black}, we discuss the case where the defender can estimate the target label $\hat{y}$ and shows that shared adversarial examples can also be beneficial for adversarial training with target label.

Given a target label $\hat{y}$, we define the Targeted Adversarial Risk as

\begin{equation}
    \gR_{tadv}(\htt) = \frac{\sum_{i=1}^N \max_{\veps_i \in \gS}\ind(\htt(\tilde{\vx}_{i, \veps_i})=\hat{y}, \vx_i\in\Dn)}{|\Dn|}.
\end{equation}

By the definition of  $\hat{y}$, we have $\gR_{bd}(\htt)\leq \gR_{tadv}(\htt) $ under Assumption 1.

Then, we have

\begin{equation}
\begin{aligned}
    &\gR_{tadv}(\htt) = \frac{1}{|\Dn|} \sum_{i=1}^N \max_{\veps_i \in \gS} \Big\{\ind(\vx_i\in \Ds)\ind(\htt(\tilde{\vx}_{i, \veps_i})=\hbd(\tilde{\vx}_{i, \veps_i})=\hat{y}) \\
    &+ \ind(\vx_i\in \Ds)\ind(\htt(\tilde{\vx}_{i, \veps_i})=\hat{y},\hbd(\tilde{\vx}_{i, \veps_i})\neq\hat{y}) + \ind(\vx_i\in \Dn\setminus\Ds, \htt(\tilde{\vx}_{i, \veps_i})=\hat{y})\Big\}.
\end{aligned}   
\end{equation}

Then, similar as $\gR_{sub}$, we define the targeted sub-adversarial risk $\gR_{tsub}$ by removing the second component in $\gR_{tadv}(\htt) $:
\begin{align*}
    \gR_{tsub}(\htt) =& \frac{1}{|\Dn|} \sum_{i=1}^N \max_{\veps_i \in \gS} \Big\{\ind(\vx_i\in \Ds)\ind(\htt(\tilde{\vx}_{i, \veps_i})=\hbd(\tilde{\vx}_{i, \veps_i})=\hat{y})\\
    &+ \ind(\vx_i\in \Dn\setminus\Ds, \htt(\tilde{\vx}_{i, \veps_i})=\hat{y})\Big\}.
\end{align*}

which considers  shared targeted adversarial risk on $\Ds$ and vanilla targeted adversarial risk on $\Dn\setminus\Ds$.

Since the removed component is irrelevant to backdoor risk, we have $\gR_{bd}\leq \gR_{tsub}$ under Assumption~1.

Therefore, we reach the following proposition
\begin{appprop}
Under Assumption~\ref{ass:add} and Assumption 1, for a classifier $\htt$, the following inequalities hold
$$\gR_{bd}(\htt)\leq \gR_{tsub}(\htt)\leq \gR_{tadv}(\htt)\leq \gR_{adv}(\htt).$$
\end{appprop}
Therefore, the shared adversarial examples also benefit adversarial training with target label.

Moreover, since $\gR_{tsub}(\htt)$ only consider shared adversarial example with target label $\hat{y}$, the following proposition holds,
\begin{appprop}
Under Assumption~\ref{ass:add} and Assumption 1, for a classifier $\htt$, the following inequalities hold
$$\gR_{bd}(\htt)\leq \gR_{tsub}(\htt)\leq \gR_{sub}(\htt)\leq \gR_{adv}(\htt).$$
\end{appprop}
The above proposition shows that  $\gR_{tsub}(\htt)$ is a tighter bound of $\gR_{bd}(\htt)$ than $\gR_{sub}(\htt)$.

\subsection{Extension to Multi-target/Multi-trigger}
In previous sections, we assume that there is only one trigger and one target for all samples. Now, we consider to a more general case where a set of triggers $\mV$ and a set of target labels $\hat{\mY}$ are given. This is a more challenging case since there may be multiple triggers and/or multiple targets for one sample.

We define $T$ to be a mapping from the sample to the corresponding set of feasible trigger-target pairs. Therefore, given a sample $\vx$, the set of all feasible trigger-target pairs is given by $T(\vx)$.  Then, Assumption 1 is extended to all triggers accordingly, \ie, $ g(\vx,\Delta)-\vx \in \gS, \forall (\Delta,\hat{y})\in T(\vx) $.

In this setting, we define that a classifier $\htt$ is said to be attacked by a poisoned sample generated by $\vx$ if there exists  $(\Delta,\hat{y})\in T(\vx)$ such that $\htt(g(\vx,\Delta))=\hat{ y}\neq y$.

Then, the (maximum) backdoor risk on $\Db$ is defined to be
\begin{equation}
    \gR_{bd}(\htt) = \frac{\sum_{i=1}^N \max_{(\Delta_i,\hat{y}_i)\in T(\vx_i))}\ind( \htt(g(\vx_i,\Delta_i))=\hat{y}_i\neq y)}{|\Db|}.
\end{equation}
Similarly, the adversarial risk  on $\Db$ is defined as
$$  \gR_{adv} = \frac{1}{|\Db|}{\sum_{i=1}^N \max_{\veps_i\in \gS}\big\{\ind( \htt(g(\vx_i,\veps_i))\neq y)\big\}}.$$

We define $\Ds = \{\vx\in \Db: \hbd(g(\vx,\Delta))=\hat{y}\neq y, \forall (\Delta,\hat{y})\in T(\vx)\}$ to be the set of samples on which planting any feasible trigger can activate the corresponding backdoor. Then, the corresponding sub-adversarial risk is defined as 
\begin{align*}
        \gR_{sub} = &\frac{1}{|\Db|}{\sum_{i=1}^N \max_{\veps_i\in \gS}\big\{\ind( \htt(g(\vx_i,\veps_i))=\hbd(g(\vx_i,\veps_i))\neq y, \vx_i\in \Ds)}\\
 & + \ind( \htt(g(\vx_i,\veps_i)) \neq y, \vx_i\in \Db\setminus\Ds)\big\}.
\end{align*}
which considers shared adversarial risk on $\Ds$ and vanilla targeted adversarial risk on $\Db\setminus\Ds$.

By definition, we have $\gR_{sub}\leq \gR_{adv}$

Moreover, we have
\begin{align*}
    \gR_{bd}(\htt) &= \frac{\sum_{i=1}^N \max_{(\Delta_i,\hat{y}_i)\in T(\vx_i))}\ind( \htt(g(\vx_i,\Delta_i))=\hat{y}_i\neq y)}{|\Db|}\\
 &= \frac{1}{|\Db|}{\sum_{i=1}^N \max_{(\Delta_i,\hat{y}_i)\in T(\vx_i))}\big\{\ind( \htt(g(\vx_i,\Delta_i))=\hbd(g(\vx_i,\Delta_i)), \hbd(g(\vx_i,\Delta_i)) = \hat{y}_i\neq y)}\\
 & + \ind( \htt(g(\vx_i,\Delta_i)) \neq y, \hbd(g(\vx_i,\Delta_i)) \neq \hat{y}_i)\big\}\\ 
  &\leq \frac{1}{|\Db|}{\sum_{i=1}^N \max_{(\Delta_i,\hat{y}_i)\in T(\vx_i))}\big\{\ind( \htt(g(\vx_i,\Delta_i))=\hbd(g(\vx_i,\Delta_i))\neq y, \vx_i\in \Ds)}\\
 & + \ind( \htt(g(\vx_i,\Delta_i)) \neq y, \vx_i\in \Db\setminus\Ds)\big\}\\ 
  &\leq \frac{1}{|\Db|}{\sum_{i=1}^N \max_{\veps_i\in \gS}\big\{\ind( \htt(g(\vx_i,\veps_i))=\hbd(g(\vx_i,\veps_i))\neq y, \vx_i\in \Ds)}\\
 & + \ind( \htt(g(\vx_i,\veps_i)) \neq y, \vx_i\in \Db\setminus\Ds)\big\}\\ 
 &=\gR_{sub}.
 \end{align*}
Therefore, for a backdoor attack with multiple targets and/or multiple triggers, the proposed sub-adversarial risk $\gR_{sub}$ is still a tighter upper bound of $\gR_{bd}$ compared to $\gR_{adv}$.

\subsection{Relaxation gap}
To build a practical objective for defending against a backdoor attack, two relaxations are applied to $\gR_{sub}$. Here, we discuss the relaxation gap in the relaxing process:
\begin{itemize}[leftmargin=*]
    \item \textbf{Replace $\Ds$ by $\Dn$}.
    
    Since $\Ds$ is not accessible, the first relaxation is to replace $\Ds$ by $\Dn$ and 
    consider shared adversarial examples on all samples in $\Dn$. Denote $\gR_1(\htt)$ to be the relaxed risk after replacing  $\Ds$ by $\Dn$. Since $\gR_{sub}$ can be decomposed as follows:
    \begin{equation*}
    \begin{aligned}
    \gR_{sub}(\htt) = &\frac{1}{|\Dn|} \sum_{i=1}^N \max_{\veps_i \in \gS} \Big\{\ind(\vx_i\in \Ds)\ind(\htt(\tilde{\vx}_{i, \veps_i})=\hbd(\tilde{\vx}_{i, \veps_i}), \tilde{\vx}_{i, \veps_i} \in \gA_{\veps_i,\vtheta}) \\
    & + \ind(\vx_i\in \Dn\setminus\Ds, \tilde{\vx}_{i, \veps_i} \in \gA_{\veps_i,\vtheta})\Big\}\\
    &=\frac{1}{|\Dn|} \sum_{i=1}^N \max_{\veps_i \in \gS} \Big\{\ind(\vx_i\in\Dn)\ind(\htt(\tilde{\vx}_{i, \veps_i})=\hbd(\tilde{\vx}_{i, \veps_i}), \tilde{\vx}_{i, \veps_i} \in \gA_{\veps_i,\vtheta}) \\
    & + \ind(\vx_i\in \Dn\setminus\Ds)\ind(\htt(\tilde{\vx}_{i, \veps_i})\neq\hbd(\tilde{\vx}_{i, \veps_i}), \tilde{\vx}_{i, \veps_i} \in \gA_{\veps_i,\vtheta})\Big\}.
    \end{aligned}   
    \end{equation*}
    We can find that the relaxation gap is 
    \begin{align*}
    \gR_{sub}-\gR_1&\leq\frac{1}{|\Dn|} \sum_{i=1}^N \max_{\veps_i \in \gS} \ind(\vx_i\in \Dn\setminus\Ds)\ind(\htt(\tilde{\vx}_{i, \veps_i})\neq\hbd(\tilde{\vx}_{i, \veps_i}), \tilde{\vx}_{i, \veps_i} \in \gA_{\veps_i,\vtheta})\\
    &\leq \frac{1}{|\Dn|} \sum_{i=1}^N \ind(\vx_i\in \Dn\setminus\Ds)\\
    & = 1-\gR_{bd}(\hbd).        
    \end{align*}
    And the relaxation gap $\text{GAP}_1$ is bounded by $1-\gR_{bd}(\hbd)$. Since a poisoned model $\hbd$ usually has a high ASR for the backdoor attack, the gap $\text{GAP}_1$ is usually negligible.

    \textbf{Remark:} We remark that the experiment results show that our method also works well for the poisoned models with low ASR. See Section \ref{app:;add} for more details.

    \item \textbf{Replace $\Dn$ by $\Db$}.  
    
    The second relaxation is to replace $\Dn$ by $\Db$ when $\hat{y}$ is not accessible. Let $\gD_{\hat{y}} = \Db\setminus\Dn $. Then, the risk after the second relaxation is:
    
    \resizebox{\linewidth}{!}{
    \begin{minipage}{\linewidth}
    \begin{equation*}
    \begin{aligned}
          &\gR_{share}(\htt) = \frac{1}{N}\sum_{i=1}^N \max_{\veps_i\in \gS} \Big\{\ind(\htt(\tilde{\vx}_{i, \veps_i})= \hbd(\tilde{\vx}_{i, \veps_i}), \tilde{\vx}_{i, \veps_i} \in \gA_{\veps_i,\vtheta})\Big\} \\
            &= \frac{|\Dn|}{N} \frac{1}{|\Dn|}\sum_{i=1}^N \max_{\veps_i\in \gS} \Big\{\ind(\vx_i\in \Dn)\ind(\htt(\tilde{\vx}_{i, \veps_i})= \hbd(\tilde{\vx}_{i, \veps_i}), \tilde{\vx}_{i, \veps_i} \in \gA_{\veps_i,\vtheta})\Big\} \\
            &+ \frac{|\gD_{\hat{y}}|}{N}\frac{1}{|\gD_{\hat{y}}|}\sum_{i=1}^N \max_{\veps_i\in \gS} \Big\{\ind(\vx_i\in \gD_{\hat{y}})\ind(\htt(\tilde{\vx}_{i, \veps_i})= \hbd(\tilde{\vx}_{i, \veps_i}), \tilde{\vx}_{i, \veps_i} \in \gA_{\veps_i,\vtheta})\Big\}.
    \end{aligned}
    \end{equation*}
    \end{minipage}}
    where  the first component is $\gR_1$ scaled by $ \frac{|\Dn|}{N}$ and the second component is the vanilla adversarial risk on $\gD_{\hat{y}}$, scaled by $\frac{|\gD_{\hat{y}}|}{N}$.
    So, the relaxation gap for the second relaxation is
    
    \resizebox{\linewidth}{!}{
    \begin{minipage}{\linewidth}
    \begin{equation*}
    \begin{aligned}
          \gR_{1}-\gR_{share} = &\frac{|\gD_{\hat{y}}|}{N}\gR_{1}
          \\& -\frac{|\gD_{\hat{y}}|}{N}\frac{1}{|\gD_{\hat{y}}|}\sum_{i=1}^N \max_{\veps_i\in \gS} \Big\{\ind(\vx_i\in \gD_{\hat{y}})\ind(\htt(\tilde{\vx}_{i, \veps_i})= \hbd(\tilde{\vx}_{i, \veps_i}), \tilde{\vx}_{i, \veps_i} \in \gA_{\veps_i,\vtheta})\Big\}.
    \end{aligned}
    \end{equation*}
    \end{minipage}}
    We can find that the generalization gap is related to the $\frac{|\gD_{\hat{y}}|}{N}$, \ie, the portion of samples whose labels are target labels, and negligible when $\frac{|\gD_{\hat{y}}|}{N}$ is small.
    
    \textbf{Remark}: For using $\gR_{share}(\htt)$ to mitigate the backdoor, the influence of  $\frac{|\Dn|}{N}$ can be eliminated by altering the trade-off parameter in Problem (8). Also, adding triggers to samples labeled $\hat{y}$ is harmless during the testing/deployment phase. So, the vanilla adversarial risk on $\gD_{\hat{y}}$ has a negligible negative impact on defending the backdoor. Therefore, replacing $\Dn$ with $\Db$ has little negative impact on mitigating the backdoor.
\end{itemize}



\section{Algorithm}
In this section, we present Algorithm~\ref{alg:SAU}, the Shared Adversarial Unlearning (SAU) algorithm for solving the proposed bi-level optimization problem and its variants.
\begin{algorithm}[H]
\caption{Shared  Adversarial Unlearning\label{alg:SAU}}
\begin{algorithmic}
\STATE \textbf{Input:} Training set $\Db$, poisoned model $\hbd$, PGD step size $\eta>0$, number of PGD steps $T_{adv}$, perturbation set $\gS$, max iteration number $T$.
\STATE Initialize $h_{\vtheta} = \hbd$.
\FOR {$t=0,...,T-1$}
\FOR {Each mini-batch in $\Db$}
\STATE Initialize perturbation $\veps$.
\FOR {$t_{adv} = 0, ..., T_{adv}-1$}
\STATE Compute gradient $\vg$ of $\veps$ w.s.t the inner maximization step in \ref{eq::overall}.
\STATE Update $\veps \to \prod_{\gS}(\veps + \eta \text{ sign}(\vg))$ where $\prod$ is the projection operation.
\ENDFOR
\STATE Update $\vtheta$ w.s.t the outer minimization step in \ref{eq::overall}.
\ENDFOR
\ENDFOR
\end{algorithmic}
\end{algorithm}

\section{Experiment details}
In our experiments, all baselines and settings are adapted from \href{https://github.com/SCLBD/BackdoorBench}{BackdoorBench} \cite{wubackdoorbench}, and we present the experiment details in this section.
\subsection{Attack details}

\begin{itemize}
    \item BadNets \cite{gu2019badnets} is one of the earliest works for backdoor learning, which inserts a small patch of fixed pattern to replace some pixels in the image. We use the default setting in BackdoorBench. 

    \item Blended backdoor attack (Blended) \cite{chen2017targeted} uses an alpha-blending strategy to fuse images with fixed patterns. We set $\alpha=0.2$ as the default in BackdoorBench. We remark that  \textbf{since such a large $\alpha$ produces visual-perceptible change to clean samples, the Blended Attack in this setting is very challenging for all defense methods. }

    \item Input-aware dynamic backdoor attack (Input-Aware)\cite{nguyen2020input} is a training-controllable attack that learns a trigger generator to produce the sample-specific trigger in the training process of the model. We use the default setting in BackdoorBench.

    \item Low-frequency attack (LF) \cite{zeng2021rethinking} uses smoothed trigger by filtering high-frequency artifacts from a UAP. We use the default setting in BackdoorBench.
    
    \item Sinusoidal signal backdoor attack (SIG) \cite{barni2019new} is a clean-label attack that uses a sinusoidal signal as the trigger to perturb the clean images in the target label.  We use the default setting in BackdoorBench.
    
    \item Sample-specific backdoor attack (SSBA) \cite{li2021invisible} uses an auto-encoder to fuse a trigger into clean samples and generate poisoned samples. We use the default setting in BackdoorBench.
    
    \item Warping-based poisoned networks (WaNet) \cite{nguyen2021wanet} is also a training-controllable attack that uses a warping function to perturb the clean samples to construct the poisoned samples. We use the default setting in BackdoorBench.
\end{itemize}

\subsection{Defense details}
The details for each defense used in our experiment are summarized below:
\begin{itemize}
    \item  ANP \cite{wu2021adversarial} is a pruning-based method that prunes neurons sensitive to weight perturbation. Note that in BackdoorBench, ANP can select its pruning threshold by grid searching on the test dataset (default way in BackdoorBench) or a given constant threshold. To produce a fair comparison with other baselines, we use a constant threshold for ANP and set the pruning threshold to $0.4$, which is found to produce better results than the recommended constant threshold of $0.2$ in BackdoorBench. All other settings are the same as the default setting in BackdoorBench.
    
    \item  FP \cite{liu2018fine} is a pruning-based method that prunes neurons according to their activations and then fine-tunes the model to keep clean accuracy. We use the default setting in BackdoorBench.
    
    \item NAD \cite{li2021neural} uses Attention Distillation to mitigate backdoors.  We use the default setting in BackdoorBench.
    
    \item  NC \cite{wang2019neural} first optimizes a possible trigger to detect whether the model is backdoored. Then, if the model is detected as a backdoored model, it mitigates
     the backdoor by unlearning the optimized trigger. We use the default setting in BackdoorBench. We remark that \textbf{if the model is detected as a clean model, NC just returns the model without any changes.}
     
    \item  EP \cite{zheng2022preactivation} uses the distribution of activation to detect the backdoor neurons and prunes them to mitigate the backdoor.   We use the default setting in BackdoorBench.

    \item i-BAU \cite{zeng2022adversarial} uses adversarial training with UAP and hyper-gradient to mitigate the backdoor.  We use the default setting in BackdoorBench. 
    
    \item SAU (Ours) uses PGD to generate shared adversarial examples and unlearns them to mitigate the backdoor. For all experiments, we set $\lambda_1=\lambda_2=\lambda_4=1$ and $\lambda_3 = 0.01$ in $(8)$. Then, we run PGD 5 steps with $L_\infty$ norm bound $0.2$  and a large step size $0.2$ to accelerate the inner maximization in ($8$).  For unlearning step, we use the same setting as i-BAU in BackdoorBench. \ie, Adam \cite{kingma2014adam} optimizer with learning rate $0.0001$. We run SAU 100 epochs in CIFAR-10 and GTSRB. In Tiny ImageNet, we run SAU 20 epochs.
    
    \item Vanilla Adversarial Training (vanilla AT) \cite{madrytowards} use PGD to generate vanilla adversarial examples and unlearns them to mitigate the backdoor. In all experiments, vanilla AT uses the same setting as SAU, including the step size, norm bound, optimizer, and learning rate, except for the experiment in Figure~$1$, where we train vanilla AT 100 epochs on Blended Attack with PreAct-ResNet18 and Tiny ImageNet to show its long-term performance. 

\end{itemize}

\paragraph{Adaptation to batch normalization } Batch normalization (BN) has been widely used in modern deep neural networks due to improved convergence. However, recent works show that it is difficult for BN to estimate the correct normalization statistics of a mixture of distributions of adversarial examples and clean examples \cite{XieY20,wang2022removing}. Such a problem is magnified when we adversarially fine-tune a model with a small set of samples and may destroy the generalization ability of the model due to biased BN statistics.  To address this problem, we employ the fixed BN strategy in the adversarial unlearning process, which has been used in implementing i-BAU \cite{zeng2022adversarial}. See i-BUA's  \href{https://github.com/YiZeng623/I-BAU}{official implementation} and \href{https://github.com/SCLBD/BackdoorBench}{BackdoorBench implementation} for more details. Note that in all experiments, the fixed BN strategy is applied to i-BAU, SAU, and vanilla AT when BN is used in the model architecture like PreAct-ResNet18. 

\paragraph{Experiments on VGG19} For VGG19 \cite{simonyan2014very}, BN is not used. Therefore, the fixed BN strategy is not applied. Moreover, some methods that depend on BN, such as ANP and EP, are not applicable to VGG19.

\section{Additional experiments}
\label{app:;add}
\subsection{Main experiments on VGG19}
\begin{table}[htbp]
\centering
\caption{Results on CIFAR-10 with VGG19 and poisoning ratio $10\%$.}
\label{cifar10_vgg19_1}
\setlength{\tabcolsep}{3pt} 
\scalebox{0.72}{
\begin{tabular}{c|cccc|cccc|cccc}
\toprule
Defense & \multicolumn{4}{c|}{No Defense}       & \multicolumn{4}{c|}{FP \cite{liu2018fine}}& \multicolumn{4}{c}{NC \cite{wang2019neural}}   \\ \midrule
Attack  & \multicolumn{1}{c}{ACC} & \multicolumn{1}{c}{ASR} & \multicolumn{1}{c}{R-ACC} & \multicolumn{1}{c|}{DER} & \multicolumn{1}{c}{ACC}         & \multicolumn{1}{c}{ASR}  & \multicolumn{1}{c}{R-ACC}& \multicolumn{1}{c|}{DER}  & \multicolumn{1}{c}{ACC}  & \multicolumn{1}{c}{ASR} & \multicolumn{1}{c}{R-ACC}& \multicolumn{1}{c}{DER}  \\
\midrule
BadNets \cite{gu2019badnets} & $89.36$& $95.93$& $3.81$&  N/A & $\textbf{89.23}$& $92.61$& $6.82$& $51.6$& $\underline{87.86}$& $\underline{1.00}$& $\textbf{88.01}$& $\textbf{96.72}$\\
Blended \cite{chen2017targeted} & $90.17$& $99.12$& $0.82$&  N/A & $\textbf{90.07}$& $99.11$& $0.82$& $49.96$& $85.92$& $\underline{1.79}$& $\textbf{74.13}$& $\underline{96.54}$\\
Input-Aware \cite{nguyen2020input} & $77.69$& $94.59$& $4.79$&  N/A & $\textbf{78.62}$& $86.77$& $11.79$& $53.91$& $\underline{77.67}$& $94.58$& $4.79$& $50.00$\\
LF \cite{zeng2021rethinking} & $88.94$& $93.93$& $5.62$&  N/A & $\textbf{88.98}$& $91.8$& $7.46$& $51.07$& $85.35$& $\underline{9.99}$& $\textbf{72.79}$& $\underline{90.18}$\\
SIG \cite{barni2019new} & $81.69$& $99.8$& $0.12$&  N/A & $84.52$& $99.93$& $0.07$& $50.0$& $81.69$& $99.80$& $0.12$& $50.00$\\
SSBA \cite{li2021invisible} & $89.48$& $91.86$& $7.29$&  N/A & $\underline{89.40}$& $89.66$& $9.22$& $51.06$& $\textbf{89.48}$& $91.86$& $7.29$& $50.00$\\
WaNet \cite{nguyen2021wanet} & $88.43$& $88.9$& $10.3$&  N/A & $89.61$& $73.39$& $24.57$& $57.76$& $88.43$& $88.89$& $10.30$& $50.01$\\
Average & $86.54$& $94.88$& $4.68$&  N/A & $\textbf{87.20}$& $90.47$& $8.68$& $52.19$& $85.20$& $55.42$& $36.78$& $69.06$\\

\toprule

\toprule
Defense & \multicolumn{4}{c|}{NAD \cite{li2021neural}}  &  \multicolumn{4}{c|}{i-BAU \cite{zeng2022adversarial}}        & \multicolumn{4}{c}{SAU (\textbf{Ours})} \\
\midrule
Attack  & \multicolumn{1}{c}{ACC} & \multicolumn{1}{c}{ASR} & \multicolumn{1}{c}{R-ACC} & \multicolumn{1}{c|}{DER} & \multicolumn{1}{c}{ACC}         & \multicolumn{1}{c}{ASR}  & \multicolumn{1}{c}{R-ACC}& \multicolumn{1}{c|}{DER}  & \multicolumn{1}{c}{ACC}  & \multicolumn{1}{c}{ASR} & \multicolumn{1}{c}{R-ACC}& \multicolumn{1}{c}{DER}  \\
\midrule
BadNets \cite{gu2019badnets} & $87.51$& $38.17$& $58.30$& $77.96$& $87.82$& $25.72$& $\underline{63.24}$& $84.34$& $86.71$& $\textbf{0.08}$& $32.77$& $\underline{96.60}$\\
Blended \cite{chen2017targeted} & $88.35$& $93.08$& $6.33$& $52.11$& $\underline{88.61}$& $59.86$& $\underline{32.31}$& $68.85$& $87.01$& $\textbf{1.32}$& $23.74$& $\textbf{97.32}$\\
Input-Aware \cite{nguyen2020input} & $75.70$& $\underline{23.36}$& $\textbf{54.71}$& $\underline{84.62}$& $72.15$& $26.22$& $42.51$& $81.41$& $75.11$& $\textbf{12.49}$& $\underline{46.19}$& $\textbf{89.76}$\\
LF \cite{zeng2021rethinking} & $87.85$& $59.38$& $35.01$& $66.73$& $\underline{87.97}$& $79.10$& $18.04$& $56.93$& $86.65$& $\textbf{6.43}$& $\underline{72.51}$& $\textbf{92.60}$\\
SIG \cite{barni2019new} & $\textbf{86.01}$& $99.18$& $0.77$& $50.31$& $\underline{85.06}$& $\underline{90.89}$& $\underline{5.27}$& $\underline{54.46}$& $84.87$& $\textbf{0.11}$& $\textbf{10.82}$& $\textbf{99.84}$\\
SSBA \cite{li2021invisible} & $87.65$& $37.54$& $54.58$& $76.24$& $88.08$& $\underline{3.74}$& $\underline{76.98}$& $\underline{93.36}$& $87.3$& $\textbf{1.66}$& $\textbf{78.70}$& $\textbf{94.01}$\\
WaNet \cite{nguyen2021wanet} & $\textbf{90.82}$& $44.93$& $51.18$& $71.98$& $\underline{90.31}$& $\underline{25.83}$& $\underline{67.87}$& $\underline{81.53}$& $87.07$& $\textbf{5.32}$& $\textbf{83.37}$& $\textbf{91.11}$\\
Average & $\underline{86.27}$& $56.52$& $37.27$& $68.56$& $85.71$& $\underline{44.48}$& $\underline{43.75}$& $\underline{74.41}$& $84.96$& $\textbf{3.92}$& $\textbf{49.73}$& $\textbf{94.46}$\\

\bottomrule
\end{tabular}}
\end{table}

This section provides additional experiment results on VGG19 with CIFAR-10 (Table~\ref{cifar10_vgg19_1}) and Tiny ImageNet (Table~\ref{tiny_vgg19_1}) to supplement Section 4. Table~\ref{cifar10_vgg19_1} shows that SAU outperforms all baselines on CIFAR-10 and VGG19, with a 40.56\% decrease of ASR compared to the second lowest ASR. Moreover, SAU achieves the highest DER in 6 of 7 attacks, demonstrating a significantly better tradeoff between accuracy and ASR. It also achieves the top-2 R-ACC in 5 of 7 attacks and the best average R-ACC, indicating its good ability to recover the prediction of poisoned samples.  Table~\ref{tiny_vgg19_1} shows that SAU achieves the best DER in 5 of 6 attacks and the top-2 lowest ASR in 5 of 6 attacks. Although SAU only achieves the second-best average ASR in Table~\ref{tiny_vgg19_1}, it has significantly higher accuracy than NC, which achieves the best average ASR.
\begin{table}[htbp]
\centering
\caption{Results on Tiny ImageNet with VGG19 and poisoning ratio $10\%$.}
\label{tiny_vgg19_1}
\setlength{\tabcolsep}{3pt} 
\scalebox{0.72}{
\begin{tabular}{c|cccc|cccc|cccc}
\toprule
Defense & \multicolumn{4}{c|}{No Defense}       & \multicolumn{4}{c|}{FP \cite{liu2018fine}}& \multicolumn{4}{c}{NC \cite{wang2019neural}}   \\ \midrule
Attack  & \multicolumn{1}{c}{ACC} & \multicolumn{1}{c}{ASR} & \multicolumn{1}{c}{R-ACC} & \multicolumn{1}{c|}{DER} & \multicolumn{1}{c}{ACC}         & \multicolumn{1}{c}{ASR}  & \multicolumn{1}{c}{R-ACC}& \multicolumn{1}{c|}{DER}  & \multicolumn{1}{c}{ACC}  & \multicolumn{1}{c}{ASR} & \multicolumn{1}{c}{R-ACC}& \multicolumn{1}{c}{DER}  \\
\midrule
BadNets \cite{gu2019badnets} & $42.26$& $99.99$& $0.0$&  N/A & $\textbf{40.70}$& $99.63$& $0.20$& $49.4$& $26.47$& $\underline{0.53}$& $\textbf{25.93}$& $\underline{91.83}$\\
Blended \cite{chen2017targeted} & $43.5$& $99.32$& $0.32$&  N/A & $\textbf{41.82}$& $98.67$& $0.71$& $49.48$& $35.93$& $\underline{0.03}$& $\textbf{18.87}$& $\underline{95.86}$\\
Input-Aware \cite{nguyen2020input} & $46.36$& $99.48$& $0.38$&  N/A & $\textbf{45.95}$& $99.02$& $0.69$& $50.03$& $38.51$& $\textbf{0.45}$& $\textbf{36.00}$& $\textbf{95.59}$\\
LF \cite{zeng2021rethinking} & $43.14$& $95.41$& $2.3$&  N/A & $\textbf{41.63}$& $85.22$& $5.82$& $54.34$& $10.55$& $\textbf{0.77}$& $5.37$& $\underline{81.02}$\\
SSBA \cite{li2021invisible} & $41.67$& $96.57$& $1.71$&  N/A & $\textbf{40.04}$& $69.27$& $\underline{9.10}$& $62.84$& $28.23$& $\textbf{0.21}$& $\textbf{22.86}$& $\underline{91.46}$\\
WaNet \cite{nguyen2021wanet} & $43.6$& $99.85$& $0.09$&  N/A & $\textbf{42.85}$& $99.31$& $0.46$& $49.9$& $8.14$& $\textbf{1.17}$& $7.13$& $\underline{81.61}$\\
Average & $43.42$& $98.44$& $0.8$&  N/A & $\textbf{42.16}$& $91.85$& $2.83$& $52.28$& $24.64$& $\textbf{0.53}$& $\textbf{19.36}$& $\underline{83.91}$\\

\toprule

\toprule
Defense & \multicolumn{4}{c|}{NAD \cite{li2021neural}}  &  \multicolumn{4}{c|}{i-BAU \cite{zeng2022adversarial}}        & \multicolumn{4}{c}{SAU (\textbf{Ours})} \\
\midrule
Attack  & \multicolumn{1}{c}{ACC} & \multicolumn{1}{c}{ASR} & \multicolumn{1}{c}{R-ACC} & \multicolumn{1}{c|}{DER} & \multicolumn{1}{c}{ACC}         & \multicolumn{1}{c}{ASR}  & \multicolumn{1}{c}{R-ACC}& \multicolumn{1}{c|}{DER}  & \multicolumn{1}{c}{ACC}  & \multicolumn{1}{c}{ASR} & \multicolumn{1}{c}{R-ACC}& \multicolumn{1}{c}{DER}  \\
\midrule
BadNets \cite{gu2019badnets} & $37.68$& $96.40$& $2.02$& $49.5$& $\underline{40.10}$& $100.0$& $0.0$& $48.92$& $38.25$& $\textbf{0.06}$& $\underline{14.32}$& $\textbf{97.96}$\\
Blended \cite{chen2017targeted} & $38.59$& $95.98$& $1.64$& $49.21$& $\underline{40.11}$& $98.88$& $0.47$& $48.52$& $38.24$& $\textbf{0.01}$& $\underline{6.10}$& $\textbf{97.02}$\\
Input-Aware \cite{nguyen2020input} & $36.14$& $\underline{2.79}$& $31.87$& $93.23$& $\underline{44.49}$& $58.66$& $18.85$& $69.47$& $42.87$& $6.94$& $\underline{35.15}$& $\underline{94.52}$\\
LF \cite{zeng2021rethinking} & $38.06$& $81.10$& $\underline{7.40}$& $54.62$& $\underline{40.23}$& $86.89$& $6.05$& $52.8$& $39.15$& $\underline{15.46}$& $\textbf{20.31}$& $\textbf{87.98}$\\
SSBA \cite{li2021invisible} & $36.27$& $84.86$& $5.78$& $53.15$& $\underline{38.93}$& $87.81$& $4.96$& $53.01$& $36.31$& $\underline{0.32}$& $8.98$& $\textbf{95.45}$\\
WaNet \cite{nguyen2021wanet} & $34.89$& $63.91$& $\underline{13.98}$& $63.61$& $\underline{39.82}$& $98.69$& $0.62$& $48.69$& $37.04$& $\underline{2.52}$& $\textbf{20.02}$& $\textbf{95.38}$\\
Average & $36.94$& $70.84$& $10.45$& $59.05$& $\underline{40.61}$& $88.49$& $5.16$& $53.06$& $38.64$& $\underline{4.22}$& $\underline{17.48}$& $\textbf{88.33}$\\

\bottomrule
\end{tabular}}
\end{table}

\subsection{Main experiments on GTSRB}
This section provides additional experiment results on GTSRB with PreAct-ResNet18 (Table~\ref{gtsrb_preactresnet18_1}) and VGG19 (Table~\ref{gtsrb_vgg19_1}) to supplement Section 4. In both tables, SAU achieves the top-2 lowest ASR in 4 of 6 attacks and the lowest ASR on average. Furthermore, SAU achieves the top-2 highest DER in 4 of 6 attacks for PreAct-ResNet18 (Table~\ref{gtsrb_preactresnet18_1}) and 5 of 6 attacks for VGG19 (Table~\ref{gtsrb_vgg19_1}).

\begin{table}[htbp]
\centering
\caption{Results on GTSRB with PreAct-ResNet18 and poisoning ratio $10\%$.}
\label{gtsrb_preactresnet18_1}
\setlength{\tabcolsep}{3pt} 
\scalebox{0.68}{
\begin{tabular}{c|cccc|cccc|cccc|cccc}
\toprule
Defense & \multicolumn{4}{c|}{No Defense}    & \multicolumn{4}{c|}{ANP \cite{wu2021adversarial}}    & \multicolumn{4}{c|}{FP \cite{liu2018fine}}& \multicolumn{4}{c}{NC \cite{wang2019neural}}   \\ \midrule
Attack  & \multicolumn{1}{c}{ACC} & \multicolumn{1}{c}{ASR} & \multicolumn{1}{c}{R-ACC} & \multicolumn{1}{c|}{DER}& \multicolumn{1}{c}{ACC}  & \multicolumn{1}{c}{ASR} & \multicolumn{1}{c}{R-ACC} & \multicolumn{1}{c|}{DER}  & \multicolumn{1}{c}{ACC}         & \multicolumn{1}{c}{ASR}  & \multicolumn{1}{c}{R-ACC}& \multicolumn{1}{c|}{DER}  & \multicolumn{1}{c}{ACC}  & \multicolumn{1}{c}{ASR} & \multicolumn{1}{c}{R-ACC}& \multicolumn{1}{c}{DER}  \\
\midrule
BadNets \cite{gu2019badnets} & $97.24$& $59.25$& $4.42$&  N/A & $96.89$& $0.06$& $96.79$& $79.42$& $\underline{98.21}$& $0.09$& $70.93$& $79.58$& $97.48$& $\underline{\textbf{0.01}}$& $97.45$& $\underline{\textbf{79.62}}$\\
Blended \cite{chen2017targeted} & $98.58$& $99.99$& $0.0$&  N/A & $\textbf{98.75}$& $99.82$& $0.18$& $50.09$& $98.38$& $100.0$& $0.0$& $49.9$& $97.76$& $\underline{8.03}$& $\textbf{59.74}$& $\underline{95.57}$\\
Input-Aware \cite{nguyen2020input} & $97.26$& $92.74$& $7.23$&  N/A & $\textbf{99.14}$& $\textbf{0.00}$& $\underline{96.09}$& $\textbf{96.37}$& $98.08$& $2.32$& $88.91$& $95.21$& $\underline{98.55}$& $0.01$& $\textbf{96.71}$& $96.36$\\
LF \cite{zeng2021rethinking} & $97.93$& $99.57$& $0.42$&  N/A & $97.80$& $81.38$& $16.45$& $59.03$& $97.59$& $99.7$& $0.27$& $49.83$& $\underline{97.97}$& $\underline{1.34}$& $\textbf{77.37}$& $\textbf{99.11}$\\
SSBA \cite{li2021invisible} & $97.98$& $99.56$& $0.49$&  N/A & $\underline{97.86}$& $98.73$& $1.13$& $50.36$& $97.75$& $99.46$& $0.45$& $49.94$& $97.72$& $\textbf{0.29}$& $\textbf{87.30}$& $\textbf{99.50}$\\
WaNet \cite{nguyen2021wanet} & $97.74$& $94.25$& $5.59$&  N/A & $98.00$& $\underline{\textbf{0.00}}$& $97.08$& $\underline{\textbf{97.12}}$& $97.62$& $88.07$& $8.71$& $53.03$& $\underline{98.25}$& $0.00$& $\textbf{98.03}$& $97.12$\\
Average & $97.79$& $90.89$& $3.02$&  N/A & $\underline{98.07}$& $46.66$& $51.29$& $68.91$& $97.94$& $64.94$& $28.21$& $61.07$& $97.96$& $\underline{1.61}$& $\textbf{86.10}$& $\underline{88.18}$\\

\toprule

\toprule
Defense & \multicolumn{4}{c|}{NAD \cite{li2021neural}}  & \multicolumn{4}{c|}{EP \cite{zheng2022preactivation}} & \multicolumn{4}{c|}{i-BAU \cite{zeng2022adversarial}}        & \multicolumn{4}{c}{SAU (\textbf{Ours})} \\
\midrule
Attack  & \multicolumn{1}{c}{ACC} & \multicolumn{1}{c}{ASR} & \multicolumn{1}{c}{R-ACC} & \multicolumn{1}{c|}{DER}  &  \multicolumn{1}{c}{ACC} & \multicolumn{1}{c}{ASR} & \multicolumn{1}{c}{R-ACC} & \multicolumn{1}{c|}{DER}  & \multicolumn{1}{c}{ACC}         & \multicolumn{1}{c}{ASR}  & \multicolumn{1}{c}{R-ACC}& \multicolumn{1}{c|}{DER}  & \multicolumn{1}{c}{ACC}  & \multicolumn{1}{c}{ASR} & \multicolumn{1}{c}{R-ACC}& \multicolumn{1}{c}{DER}  \\
\midrule
BadNets \cite{gu2019badnets} & $\textbf{98.69}$& $0.63$& $\textbf{98.00}$& $79.31$& $97.69$& $10.94$& $88.22$& $74.16$& $96.03$& $0.02$& $96.07$& $79.01$& $98.0$& $0.01$& $\underline{97.92}$& $79.62$\\
Blended \cite{chen2017targeted} & $\underline{98.61}$& $100.0$& $0.00$& $50.0$& $97.9$& $100.0$& $0.0$& $49.66$& $95.65$& $86.45$& $6.52$& $55.31$& $96.73$& $\textbf{3.41}$& $\underline{34.42}$& $\textbf{97.36}$\\
Input-Aware \cite{nguyen2020input} & $98.27$& $40.65$& $58.68$& $76.04$& $98.5$& $49.49$& $49.5$& $71.62$& $97.93$& $1.94$& $93.58$& $95.4$& $98.09$& $\underline{0.01}$& $93.02$& $\underline{96.36}$\\
LF \cite{zeng2021rethinking} & $\textbf{98.14}$& $51.83$& $\underline{40.02}$& $73.87$& $97.59$& $99.2$& $0.72$& $50.02$& $95.59$& $17.3$& $32.16$& $89.97$& $96.08$& $\textbf{0.37}$& $16.89$& $\underline{98.68}$\\
SSBA \cite{li2021invisible} & $\textbf{97.95}$& $99.39$& $0.59$& $50.07$& $97.66$& $98.66$& $1.13$& $50.29$& $96.13$& $1.27$& $\underline{58.10}$& $98.22$& $96.94$& $\underline{0.45}$& $40.33$& $\underline{99.04}$\\
WaNet \cite{nguyen2021wanet} & $\textbf{98.61}$& $0.56$& $\underline{97.63}$& $96.84$& $96.96$& $8.97$& $82.93$& $92.25$& $96.92$& $0.11$& $96.25$& $96.66$& $97.86$& $0.02$& $97.42$& $97.11$\\
Average & $\textbf{98.38}$& $48.84$& $49.15$& $68.02$& $97.72$& $61.21$& $37.08$& $62.57$& $96.38$& $17.85$& $\underline{63.78}$& $80.65$& $97.28$& $\textbf{0.71}$& $63.33$& $\textbf{88.31}$\\

\bottomrule
\end{tabular}}
\end{table}

\begin{table}[h]
\centering
\caption{Results on GTSRB with VGG19 and poisoning ratio $10\%$.}
\label{gtsrb_vgg19_1}
\setlength{\tabcolsep}{3pt} 
\scalebox{0.72}{
\begin{tabular}{c|cccc|cccc|cccc}
\toprule
Defense & \multicolumn{4}{c|}{No Defense}       & \multicolumn{4}{c|}{FP \cite{liu2018fine}}& \multicolumn{4}{c}{NC \cite{wang2019neural}}   \\ \midrule
Attack  & \multicolumn{1}{c}{ACC} & \multicolumn{1}{c}{ASR} & \multicolumn{1}{c}{R-ACC} & \multicolumn{1}{c|}{DER} & \multicolumn{1}{c}{ACC}         & \multicolumn{1}{c}{ASR}  & \multicolumn{1}{c}{R-ACC}& \multicolumn{1}{c|}{DER}  & \multicolumn{1}{c}{ACC}  & \multicolumn{1}{c}{ASR} & \multicolumn{1}{c}{R-ACC}& \multicolumn{1}{c}{DER}  \\
\midrule
BadNets \cite{gu2019badnets} & $96.25$& $57.11$& $4.88$&  N/A & $\textbf{97.02}$& $2.89$& $18.64$& $77.11$& $\underline{96.86}$& $\textbf{0.00}$& $\textbf{96.64}$& $\textbf{78.56}$\\
Blended \cite{chen2017targeted} & $95.98$& $99.86$& $0.04$&  N/A & $\underline{96.48}$& $99.79$& $0.08$& $50.03$& $95.32$& $20.80$& $\textbf{38.37}$& $89.20$\\
Input-Aware \cite{nguyen2020input} & $96.03$& $76.69$& $22.35$&  N/A & $\underline{97.08}$& $24.09$& $\underline{73.09}$& $76.3$& $96.71$& $\underline{0.02}$& $\textbf{94.22}$& $\underline{88.33}$\\
LF \cite{zeng2021rethinking} & $95.05$& $98.79$& $0.99$&  N/A & $\underline{95.28}$& $98.13$& $1.48$& $50.33$& $93.25$& $\underline{0.60}$& $\textbf{67.61}$& $98.19$\\
SSBA \cite{li2021invisible} & $96.43$& $99.31$& $0.43$&  N/A & $\underline{96.45}$& $99.0$& $0.64$& $50.16$& $\textbf{96.56}$& $\textbf{1.03}$& $\textbf{79.49}$& $\textbf{99.14}$\\
WaNet \cite{nguyen2021wanet} & $95.27$& $92.09$& $7.35$&  N/A & $96.90$& $68.48$& $29.00$& $61.81$& $96.84$& $\textbf{1.11}$& $\textbf{94.67}$& $\textbf{95.49}$\\
Average & $95.84$& $87.31$& $6.01$&  N/A & $\underline{96.54}$& $65.4$& $20.49$& $59.39$& $95.92$& $3.93$& $\textbf{78.50}$& $85.56$\\

\toprule

\toprule
Defense & \multicolumn{4}{c|}{NAD \cite{li2021neural}}  &  \multicolumn{4}{c|}{i-BAU \cite{zeng2022adversarial}}        & \multicolumn{4}{c}{SAU (\textbf{Ours})} \\
\midrule
Attack  & \multicolumn{1}{c}{ACC} & \multicolumn{1}{c}{ASR} & \multicolumn{1}{c}{R-ACC} & \multicolumn{1}{c|}{DER} & \multicolumn{1}{c}{ACC}         & \multicolumn{1}{c}{ASR}  & \multicolumn{1}{c}{R-ACC}& \multicolumn{1}{c|}{DER}  & \multicolumn{1}{c}{ACC}  & \multicolumn{1}{c}{ASR} & \multicolumn{1}{c}{R-ACC}& \multicolumn{1}{c}{DER}  \\
\midrule
BadNets \cite{gu2019badnets} & $96.78$& $73.52$& $25.73$& $50.0$& $95.81$& $\underline{0.01}$& $\underline{94.70}$& $78.33$& $96.21$& $0.02$& $79.69$& $\underline{78.53}$\\
Blended \cite{chen2017targeted} & $\textbf{96.60}$& $98.27$& $0.55$& $50.79$& $95.11$& $\textbf{0.79}$& $\underline{16.55}$& $\textbf{99.10}$& $95.45$& $\underline{1.79}$& $12.61$& $\underline{98.77}$\\
Input-Aware \cite{nguyen2020input} & $\textbf{97.77}$& $53.03$& $45.94$& $61.83$& $96.92$& $10.06$& $72.08$& $83.31$& $96.48$& $\textbf{0.01}$& $11.34$& $\textbf{88.34}$\\
LF \cite{zeng2021rethinking} & $94.69$& $97.07$& $2.43$& $50.68$& $\textbf{95.51}$& $0.83$& $\underline{22.43}$& $\underline{98.98}$& $95.17$& $\textbf{0.02}$& $15.83$& $\textbf{99.38}$\\
SSBA \cite{li2021invisible} & $96.38$& $98.38$& $0.95$& $50.44$& $94.57$& $\underline{2.55}$& $\underline{61.79}$& $\underline{97.45}$& $95.92$& $7.44$& $31.49$& $95.68$\\
WaNet \cite{nguyen2021wanet} & $\textbf{97.47}$& $39.34$& $57.96$& $76.38$& $96.86$& $6.28$& $81.71$& $92.90$& $\underline{97.04}$& $\underline{5.89}$& $\underline{88.15}$& $\underline{93.10}$\\
Average & $\textbf{96.62}$& $76.6$& $22.26$& $55.73$& $95.80$& $\underline{3.42}$& $\underline{58.21}$& $\underline{85.72}$& $96.04$& $\textbf{2.53}$& $39.85$& $\textbf{86.26}$\\

\bottomrule
\end{tabular}}
\end{table}

\subsection{Experiments with $5\%$ poisoning ratio}
This section reports the experiment results for comparing SAU with baselines with a poisoning ratio of 5\% on CIFAR-10 and Tiny-ImageNet. As with the results for a poisoning ratio of 10\%, SAU achieves the top-2 lowest ASR in most of the attacks. Tables~\ref{cifar10_preactresnet18_05}, \ref{cifar10_vgg19_05} and \ref{tiny_preactresnet18_05} show that SAU achieves the lowest average ASR. Table~\ref{tiny_vgg19_05} shows that SAU achieves the second lowest average ASR with a significantly higher average accuracy (+10.41\%) than NC, which achieves the best ASR. In summary, SAU still outperforms other baselines for a poisoning ratio of 5

\begin{table}[h]
\centering
\caption{Results on CIFAR-10 with PreAct-ResNet18 and poisoning ratio $5\%$.}
\label{cifar10_preactresnet18_05}
\setlength{\tabcolsep}{3pt} 
\scalebox{0.68}{
\begin{tabular}{c|cccc|cccc|cccc|cccc}
\toprule
Defense & \multicolumn{4}{c|}{No Defense}    & \multicolumn{4}{c|}{ANP \cite{wu2021adversarial}}    & \multicolumn{4}{c|}{FP \cite{liu2018fine}}& \multicolumn{4}{c}{NC \cite{wang2019neural}}   \\ \midrule
Attack  & \multicolumn{1}{c}{ACC} & \multicolumn{1}{c}{ASR} & \multicolumn{1}{c}{R-ACC} & \multicolumn{1}{c|}{DER}& \multicolumn{1}{c}{ACC}  & \multicolumn{1}{c}{ASR} & \multicolumn{1}{c}{R-ACC} & \multicolumn{1}{c|}{DER}  & \multicolumn{1}{c}{ACC}         & \multicolumn{1}{c}{ASR}  & \multicolumn{1}{c}{R-ACC}& \multicolumn{1}{c|}{DER}  & \multicolumn{1}{c}{ACC}  & \multicolumn{1}{c}{ASR} & \multicolumn{1}{c}{R-ACC}& \multicolumn{1}{c}{DER}  \\
\midrule
BadNets \cite{gu2019badnets} & $92.64$& $88.74$& $10.78$&  N/A & $\underline{92.38}$& $3.10$& $90.13$& $\underline{92.69}$& $\textbf{92.47}$& $17.47$& $77.9$& $85.55$& $89.89$& $\underline{1.17}$& $89.70$& $92.41$\\
Blended \cite{chen2017targeted} & $93.67$& $99.61$& $0.39$&  N/A & $93.32$& $98.00$& $1.92$& $50.63$& $\underline{93.46}$& $98.87$& $1.11$& $50.27$& $\textbf{93.66}$& $99.61$& $0.39$& $50.00$\\
Input-Aware \cite{nguyen2020input} & $91.52$& $90.2$& $8.91$&  N/A & $91.64$& $\textbf{1.08}$& $\textbf{87.01}$& $\textbf{94.56}$& $\underline{91.89}$& $52.74$& $44.49$& $68.73$& $91.51$& $90.20$& $8.91$& $50.00$\\
LF \cite{zeng2021rethinking} & $93.35$& $98.03$& $1.86$&  N/A & $92.60$& $96.98$& $2.87$& $50.15$& $\underline{93.14}$& $98.37$& $1.53$& $49.9$& $\textbf{93.35}$& $98.03$& $1.86$& $50.00$\\
SIG \cite{barni2019new} & $93.64$& $97.09$& $2.9$&  N/A & $93.09$& $92.38$& $7.39$& $52.08$& $\underline{93.27}$& $99.79$& $0.2$& $49.81$& $\textbf{93.65}$& $97.09$& $2.90$& $50.00$\\
SSBA \cite{li2021invisible} & $93.27$& $94.91$& $4.82$&  N/A & $\textbf{93.13}$& $52.66$& $43.68$& $71.06$& $\underline{93.05}$& $87.14$& $12.09$& $53.77$& $91.04$& $\textbf{0.80}$& $\textbf{87.93}$& $\textbf{95.94}$\\
WaNet \cite{nguyen2021wanet} & $91.76$& $85.5$& $13.49$&  N/A & $90.67$& $\textbf{2.31}$& $\underline{88.31}$& $\textbf{91.05}$& $\underline{92.14}$& $26.1$& $63.5$& $79.7$& $91.76$& $85.50$& $13.49$& $50.00$\\
Average & $92.84$& $93.44$& $6.16$&  N/A & $92.40$& $49.50$& $45.90$& $71.75$& $\textbf{92.77}$& $68.64$& $28.69$& $62.53$& $92.12$& $67.49$& $29.31$& $62.62$\\

\toprule

\toprule
Defense & \multicolumn{4}{c|}{NAD \cite{li2021neural}}  & \multicolumn{4}{c|}{EP \cite{zheng2022preactivation}} & \multicolumn{4}{c|}{i-BAU \cite{zeng2022adversarial}}        & \multicolumn{4}{c}{SAU (\textbf{Ours})} \\
\midrule
Attack  & \multicolumn{1}{c}{ACC} & \multicolumn{1}{c}{ASR} & \multicolumn{1}{c}{R-ACC} & \multicolumn{1}{c|}{DER}  &  \multicolumn{1}{c}{ACC} & \multicolumn{1}{c}{ASR} & \multicolumn{1}{c}{R-ACC} & \multicolumn{1}{c|}{DER}  & \multicolumn{1}{c}{ACC}         & \multicolumn{1}{c}{ASR}  & \multicolumn{1}{c}{R-ACC}& \multicolumn{1}{c|}{DER}  & \multicolumn{1}{c}{ACC}  & \multicolumn{1}{c}{ASR} & \multicolumn{1}{c}{R-ACC}& \multicolumn{1}{c}{DER}  \\
\midrule
BadNets \cite{gu2019badnets} & $91.03$& $4.73$& $87.93$& $91.2$& $91.04$& $\textbf{0.98}$& $\textbf{91.26}$& $\textbf{93.08}$& $88.8$& $2.51$& $87.82$& $91.20$& $90.54$& $1.36$& $\underline{90.32}$& $92.64$\\
Blended \cite{chen2017targeted} & $93.10$& $99.06$& $0.93$& $49.99$& $92.07$& $98.88$& $1.01$& $49.57$& $88.39$& $\underline{35.89}$& $\underline{43.28}$& $\underline{79.22}$& $91.25$& $\textbf{5.64}$& $\textbf{67.56}$& $\textbf{95.77}$\\
Input-Aware \cite{nguyen2020input} & $\textbf{93.07}$& $97.12$& $2.68$& $50.0$& $90.59$& $\underline{1.51}$& $\underline{85.09}$& $93.88$& $91.34$& $7.63$& $80.10$& $91.19$& $91.12$& $2.03$& $82.11$& $\underline{93.88}$\\
LF \cite{zeng2021rethinking} & $92.98$& $94.23$& $5.3$& $51.72$& $91.65$& $83.71$& $15.22$& $56.31$& $87.25$& $\underline{34.61}$& $\underline{41.31}$& $\underline{78.66}$& $91.31$& $\textbf{2.59}$& $\textbf{78.93}$& $\textbf{96.70}$\\
SIG \cite{barni2019new} & $92.49$& $96.98$& $2.93$& $49.48$& $92.4$& $\underline{1.86}$& $\textbf{60.29}$& $\textbf{97.00}$& $86.44$& $5.73$& $48.86$& $92.08$& $90.87$& $\textbf{0.97}$& $\underline{54.98}$& $\underline{96.68}$\\
SSBA \cite{li2021invisible} & $92.49$& $88.63$& $10.58$& $52.75$& $92.29$& $12.14$& $78.70$& $90.89$& $87.15$& $3.86$& $78.57$& $92.47$& $91.15$& $\underline{1.43}$& $\underline{85.87}$& $\underline{95.68}$\\
WaNet \cite{nguyen2021wanet} & $\textbf{93.31}$& $50.4$& $46.46$& $67.55$& $91.36$& $31.98$& $62.87$& $76.56$& $89.71$& $4.24$& $83.68$& $89.60$& $91.69$& $\underline{3.58}$& $\textbf{88.33}$& $\underline{90.93}$\\
Average & $\underline{92.64}$& $75.88$& $22.4$& $58.96$& $91.63$& $33.01$& $56.35$& $79.61$& $88.44$& $\underline{13.50}$& $\underline{66.23}$& $\underline{87.77}$& $91.13$& $\textbf{2.51}$& $\textbf{78.30}$& $\textbf{94.61}$\\

\bottomrule
\end{tabular}}
\end{table}

\begin{table}[h]
\centering
\caption{Results on CIFAR-10 with VGG19 and poisoning ratio $5\%$.}
\label{cifar10_vgg19_05}
\setlength{\tabcolsep}{3pt} 
\scalebox{0.72}{
\begin{tabular}{c|cccc|cccc|cccc}
\toprule
Defense & \multicolumn{4}{c|}{No Defense}       & \multicolumn{4}{c|}{FP \cite{liu2018fine}}& \multicolumn{4}{c}{NC \cite{wang2019neural}}   \\ \midrule
Attack  & \multicolumn{1}{c}{ACC} & \multicolumn{1}{c}{ASR} & \multicolumn{1}{c}{R-ACC} & \multicolumn{1}{c|}{DER} & \multicolumn{1}{c}{ACC}         & \multicolumn{1}{c}{ASR}  & \multicolumn{1}{c}{R-ACC}& \multicolumn{1}{c|}{DER}  & \multicolumn{1}{c}{ACC}  & \multicolumn{1}{c}{ASR} & \multicolumn{1}{c}{R-ACC}& \multicolumn{1}{c}{DER}  \\
\midrule
BadNets \cite{gu2019badnets} & $89.83$& $94.36$& $5.39$&  N/A & $\textbf{89.81}$& $90.32$& $9.09$& $52.01$& $\underline{88.51}$& $\underline{1.04}$& $\textbf{88.31}$& $\underline{96.00}$\\
Blended \cite{chen2017targeted} & $90.41$& $97.98$& $1.79$&  N/A & $\textbf{90.21}$& $97.93$& $1.84$& $49.92$& $86.92$& $\textbf{2.61}$& $\textbf{73.99}$& $\textbf{95.94}$\\
Input-Aware \cite{nguyen2020input} & $89.34$& $71.66$& $25.77$&  N/A & $\underline{89.90}$& $52.58$& $\underline{43.51}$& $59.54$& $89.58$& $\underline{2.88}$& $\textbf{85.30}$& $\underline{84.39}$\\
LF \cite{zeng2021rethinking} & $90.13$& $86.04$& $12.5$&  N/A & $\textbf{89.86}$& $83.69$& $14.77$& $51.04$& $87.25$& $\textbf{8.58}$& $\textbf{76.39}$& $\textbf{87.29}$\\
SIG \cite{barni2019new} & $90.41$& $96.76$& $2.84$&  N/A & $\textbf{90.55}$& $97.72$& $2.12$& $50.00$& $\underline{90.41}$& $96.76$& $2.84$& $50.00$\\
SSBA \cite{li2021invisible} & $89.93$& $81.19$& $17.14$&  N/A & $\textbf{89.81}$& $78.61$& $19.68$& $51.23$& $87.31$& $\textbf{2.27}$& $\textbf{83.27}$& $\textbf{88.15}$\\
WaNet \cite{nguyen2021wanet} & $89.25$& $3.96$& $86.71$&  N/A & $\underline{90.25}$& $\underline{1.66}$& $89.08$& $\underline{51.15}$& $89.25$& $3.96$& $86.69$& $50.00$\\
Average & $89.9$& $75.99$& $21.73$&  N/A & $\textbf{90.06}$& $71.79$& $25.73$& $52.13$& $88.46$& $\underline{16.87}$& $\textbf{70.97}$& $\underline{78.82}$\\

\toprule

\toprule
Defense & \multicolumn{4}{c|}{NAD \cite{li2021neural}}  &  \multicolumn{4}{c|}{i-BAU \cite{zeng2022adversarial}}        & \multicolumn{4}{c}{SAU (\textbf{Ours})} \\
\midrule
Attack  & \multicolumn{1}{c}{ACC} & \multicolumn{1}{c}{ASR} & \multicolumn{1}{c}{R-ACC} & \multicolumn{1}{c|}{DER} & \multicolumn{1}{c}{ACC}         & \multicolumn{1}{c}{ASR}  & \multicolumn{1}{c}{R-ACC}& \multicolumn{1}{c|}{DER}  & \multicolumn{1}{c}{ACC}  & \multicolumn{1}{c}{ASR} & \multicolumn{1}{c}{R-ACC}& \multicolumn{1}{c}{DER}  \\
\midrule
BadNets \cite{gu2019badnets} & $88.42$& $74.51$& $23.89$& $59.22$& $88.09$& $70.19$& $27.22$& $61.21$& $88.14$& $\textbf{0.56}$& $\underline{72.36}$& $\textbf{96.06}$\\
Blended \cite{chen2017targeted} & $\underline{88.84}$& $92.68$& $6.70$& $51.86$& $88.05$& $62.08$& $21.07$& $66.77$& $88.19$& $\underline{5.18}$& $\underline{30.78}$& $\underline{95.29}$\\
Input-Aware \cite{nguyen2020input} & $\textbf{90.51}$& $81.62$& $17.73$& $50.00$& $89.08$& $72.71$& $18.96$& $49.87$& $87.79$& $\textbf{0.10}$& $13.71$& $\textbf{85.00}$\\
LF \cite{zeng2021rethinking} & $88.82$& $54.51$& $40.07$& $65.11$& $\underline{88.97}$& $66.98$& $29.04$& $58.95$& $87.33$& $\underline{9.76}$& $\underline{70.19}$& $\underline{86.74}$\\
SIG \cite{barni2019new} & $88.96$& $\underline{91.34}$& $\underline{7.30}$& $\underline{51.98}$& $89.30$& $95.20$& $4.20$& $50.22$& $87.46$& $\textbf{73.47}$& $\textbf{14.40}$& $\textbf{60.17}$\\
SSBA \cite{li2021invisible} & $88.07$& $43.93$& $48.86$& $67.70$& $\underline{88.98}$& $4.43$& $\underline{82.91}$& $\underline{87.90}$& $86.87$& $\underline{3.48}$& $75.31$& $87.33$\\
WaNet \cite{nguyen2021wanet} & $\textbf{90.61}$& $1.67$& $\textbf{89.40}$& $51.14$& $90.00$& $\textbf{1.26}$& $\underline{89.08}$& $\textbf{51.35}$& $89.02$& $2.67$& $87.44$& $50.53$\\
Average & $\underline{89.18}$& $62.89$& $33.42$& $56.72$& $88.92$& $53.26$& $38.93$& $60.90$& $87.83$& $\textbf{13.60}$& $\underline{52.03}$& $\textbf{80.16}$\\

\bottomrule
\end{tabular}}
\end{table}

\begin{table}[t]
\centering
\caption{Results on Tiny ImageNet with PreAct-ResNet18 and poisoning ratio $5\%$.}
\label{tiny_preactresnet18_05}
\setlength{\tabcolsep}{3pt} 
\scalebox{0.68}{
\begin{tabular}{c|cccc|cccc|cccc|cccc}
\toprule
Defense & \multicolumn{4}{c|}{No Defense}    & \multicolumn{4}{c|}{ANP \cite{wu2021adversarial}}    & \multicolumn{4}{c|}{FP \cite{liu2018fine}}& \multicolumn{4}{c}{NC \cite{wang2019neural}}   \\ \midrule
Attack  & \multicolumn{1}{c}{ACC} & \multicolumn{1}{c}{ASR} & \multicolumn{1}{c}{R-ACC} & \multicolumn{1}{c|}{DER}& \multicolumn{1}{c}{ACC}  & \multicolumn{1}{c}{ASR} & \multicolumn{1}{c}{R-ACC} & \multicolumn{1}{c|}{DER}  & \multicolumn{1}{c}{ACC}         & \multicolumn{1}{c}{ASR}  & \multicolumn{1}{c}{R-ACC}& \multicolumn{1}{c|}{DER}  & \multicolumn{1}{c}{ACC}  & \multicolumn{1}{c}{ASR} & \multicolumn{1}{c}{R-ACC}& \multicolumn{1}{c}{DER}  \\
\midrule
BadNets \cite{gu2019badnets} & $56.36$& $99.86$& $0.13$&  N/A & $44.54$& $\textbf{0.02}$& $43.80$& $94.01$& $\underline{53.00}$& $96.83$& $2.86$& $49.83$& $52.97$& $65.94$& $25.45$& $65.26$\\
Blended \cite{chen2017targeted} & $56.69$& $98.75$& $0.94$&  N/A & $45.19$& $92.89$& $3.67$& $47.18$& $52.96$& $94.47$& $3.21$& $50.28$& $\textbf{53.63}$& $96.52$& $2.21$& $49.59$\\
Input-Aware \cite{nguyen2020input} & $57.87$& $98.25$& $1.51$&  N/A & $49.38$& $\underline{0.17}$& $48.84$& $94.80$& $56.27$& $89.87$& $6.52$& $53.39$& $\underline{56.48}$& $0.49$& $\textbf{53.17}$& $\underline{98.18}$\\
LF \cite{zeng2021rethinking} & $56.67$& $95.99$& $2.75$&  N/A & $41.26$& $\textbf{38.48}$& $\underline{17.21}$& $\underline{71.05}$& $\underline{53.11}$& $86.8$& $6.47$& $52.81$& $51.49$& $67.51$& $16.57$& $61.65$\\
SSBA \cite{li2021invisible} & $56.86$& $95.69$& $3.05$&  N/A & $44.93$& $50.44$& $20.23$& $66.66$& $53.26$& $79.97$& $10.63$& $56.06$& $\textbf{54.04}$& $\underline{0.04}$& $\underline{38.80}$& $\textbf{96.41}$\\
WaNet \cite{nguyen2021wanet} & $56.83$& $98.36$& $1.21$&  N/A & $29.42$& $\textbf{0.03}$& $27.29$& $85.46$& $\textbf{54.18}$& $93.36$& $2.09$& $51.18$& $53.02$& $0.27$& $\underline{50.56}$& $\underline{97.14}$\\
Average & $56.88$& $97.82$& $1.6$&  N/A & $42.45$& $\underline{30.34}$& $26.84$& $72.74$& $\underline{53.80}$& $90.22$& $5.3$& $51.94$& $53.60$& $38.46$& $31.13$& $74.03$\\

\toprule

\toprule
Defense & \multicolumn{4}{c|}{NAD \cite{li2021neural}}  & \multicolumn{4}{c|}{EP \cite{zheng2022preactivation}} & \multicolumn{4}{c|}{i-BAU \cite{zeng2022adversarial}}        & \multicolumn{4}{c}{SAU (\textbf{Ours})} \\
\midrule
Attack  & \multicolumn{1}{c}{ACC} & \multicolumn{1}{c}{ASR} & \multicolumn{1}{c}{R-ACC} & \multicolumn{1}{c|}{DER}  &  \multicolumn{1}{c}{ACC} & \multicolumn{1}{c}{ASR} & \multicolumn{1}{c}{R-ACC} & \multicolumn{1}{c|}{DER}  & \multicolumn{1}{c}{ACC}         & \multicolumn{1}{c}{ASR}  & \multicolumn{1}{c}{R-ACC}& \multicolumn{1}{c|}{DER}  & \multicolumn{1}{c}{ACC}  & \multicolumn{1}{c}{ASR} & \multicolumn{1}{c}{R-ACC}& \multicolumn{1}{c}{DER}  \\
\midrule
BadNets \cite{gu2019badnets} & $45.51$& $34.63$& $34.01$& $77.19$& $51.58$& $\underline{0.10}$& $\textbf{51.06}$& $\textbf{97.49}$& $\textbf{53.60}$& $96.58$& $3.17$& $50.26$& $51.94$& $0.78$& $\underline{50.78}$& $\underline{97.33}$\\
Blended \cite{chen2017targeted} & $46.72$& $88.89$& $4.96$& $49.94$& $\underline{53.13}$& $\underline{82.26}$& $\underline{8.58}$& $\underline{56.47}$& $52.76$& $92.01$& $4.7$& $51.41$& $52.47$& $\textbf{3.40}$& $\textbf{24.90}$& $\textbf{95.57}$\\
Input-Aware \cite{nguyen2020input} & $48.72$& $1.04$& $45.83$& $94.03$& $\textbf{57.09}$& $0.92$& $\underline{51.27}$& $\textbf{98.27}$& $56.13$& $36.99$& $40.9$& $79.76$& $53.92$& $\textbf{0.11}$& $50.89$& $97.10$\\
LF \cite{zeng2021rethinking} & $46.93$& $63.9$& $14.66$& $61.18$& $\textbf{54.04}$& $83.80$& $10.06$& $54.78$& $52.45$& $90.81$& $5.77$& $50.48$& $52.57$& $\underline{44.27}$& $\textbf{23.78}$& $\textbf{73.81}$\\
SSBA \cite{li2021invisible} & $46.65$& $42.45$& $24.65$& $71.51$& $\underline{53.87}$& $52.07$& $23.44$& $70.31$& $52.65$& $90.72$& $5.95$& $50.38$& $53.57$& $\textbf{0.02}$& $\textbf{39.77}$& $\underline{96.19}$\\
WaNet \cite{nguyen2021wanet} & $46.81$& $2.44$& $39.54$& $92.95$& $\underline{53.98}$& $\underline{0.12}$& $\textbf{52.62}$& $\textbf{97.70}$& $52.99$& $93.43$& $4.93$& $50.55$& $50.15$& $81.75$& $12.37$& $54.97$\\
Average & $46.89$& $38.89$& $27.27$& $70.97$& $\textbf{53.95}$& $36.54$& $\underline{32.84}$& $\underline{75.00}$& $53.43$& $83.42$& $10.9$& $54.69$& $52.44$& $\textbf{21.72}$& $\textbf{33.75}$& $\textbf{80.71}$\\

\bottomrule
\end{tabular}}
\end{table}

\begin{table}[htbp]
\centering
\caption{Results on Tiny ImageNet with VGG19 and poisoning ratio $5\%$.}
\label{tiny_vgg19_05}
\setlength{\tabcolsep}{3pt} 
\scalebox{0.72}{
\begin{tabular}{c|cccc|cccc|cccc}
\toprule
Defense & \multicolumn{4}{c|}{No Defense}       & \multicolumn{4}{c|}{FP \cite{liu2018fine}}& \multicolumn{4}{c}{NC \cite{wang2019neural}}   \\ \midrule
Attack  & \multicolumn{1}{c}{ACC} & \multicolumn{1}{c}{ASR} & \multicolumn{1}{c}{R-ACC} & \multicolumn{1}{c|}{DER} & \multicolumn{1}{c}{ACC}         & \multicolumn{1}{c}{ASR}  & \multicolumn{1}{c}{R-ACC}& \multicolumn{1}{c|}{DER}  & \multicolumn{1}{c}{ACC}  & \multicolumn{1}{c}{ASR} & \multicolumn{1}{c}{R-ACC}& \multicolumn{1}{c}{DER}  \\
\midrule
BadNets \cite{gu2019badnets} & $45.4$& $99.98$& $0.01$&  N/A & $\textbf{43.76}$& $94.16$& $1.66$& $52.09$& $5.05$& $\underline{0.42}$& $\underline{4.64}$& $\underline{79.60}$\\
Blended \cite{chen2017targeted} & $44.97$& $98.64$& $0.74$&  N/A & $\textbf{43.39}$& $97.45$& $1.13$& $49.81$& $36.59$& $\underline{0.09}$& $\textbf{19.99}$& $\underline{95.09}$\\
Input-Aware \cite{nguyen2020input} & $48.55$& $98.67$& $0.97$&  N/A & $\textbf{47.60}$& $98.11$& $1.44$& $49.81$& $14.76$& $\underline{0.80}$& $11.85$& $82.04$\\
LF \cite{zeng2021rethinking} & $44.04$& $91.12$& $4.31$&  N/A & $\textbf{42.08}$& $84.02$& $6.9$& $52.57$& $29.0$& $\textbf{0.60}$& $\textbf{21.02}$& $\textbf{87.74}$\\
SSBA \cite{li2021invisible} & $44.0$& $94.25$& $3.06$&  N/A & $\textbf{42.34}$& $87.24$& $5.6$& $52.68$& $4.03$& $\underline{0.35}$& $3.93$& $\underline{76.96}$\\
WaNet \cite{nguyen2021wanet} & $45.04$& $98.98$& $0.64$&  N/A & $\textbf{44.20}$& $79.4$& $8.1$& $59.37$& $31.15$& $\textbf{0.97}$& $\textbf{28.15}$& $\underline{92.06}$\\
Average & $45.33$& $96.94$& $1.62$&  N/A & $\textbf{43.90}$& $90.06$& $4.14$& $52.33$& $20.1$& $\textbf{0.54}$& $\textbf{14.93}$& $\underline{80.50}$\\

\toprule

\toprule
Defense & \multicolumn{4}{c|}{NAD \cite{li2021neural}}  &  \multicolumn{4}{c|}{i-BAU \cite{zeng2022adversarial}}        & \multicolumn{4}{c}{SAU (\textbf{Ours})} \\
\midrule
Attack  & \multicolumn{1}{c}{ACC} & \multicolumn{1}{c}{ASR} & \multicolumn{1}{c}{R-ACC} & \multicolumn{1}{c|}{DER} & \multicolumn{1}{c}{ACC}         & \multicolumn{1}{c}{ASR}  & \multicolumn{1}{c}{R-ACC}& \multicolumn{1}{c|}{DER}  & \multicolumn{1}{c}{ACC}  & \multicolumn{1}{c}{ASR} & \multicolumn{1}{c}{R-ACC}& \multicolumn{1}{c}{DER}  \\
\midrule
BadNets \cite{gu2019badnets} & $40.98$& $95.55$& $2.33$& $50.01$& $\underline{41.78}$& $99.9$& $0.06$& $48.23$& $39.95$& $\textbf{0.00}$& $\textbf{6.27}$& $\textbf{97.26}$\\
Blended \cite{chen2017targeted} & $40.51$& $94.14$& $2.50$& $50.02$& $\underline{42.17}$& $95.89$& $1.74$& $49.98$& $39.47$& $\textbf{0.01}$& $\underline{5.14}$& $\textbf{96.57}$\\
Input-Aware \cite{nguyen2020input} & $41.02$& $3.9$& $\textbf{36.31}$& $\underline{93.62}$& $\underline{44.87}$& $84.46$& $9.47$& $55.27$& $41.75$& $\textbf{0.06}$& $\underline{35.53}$& $\textbf{95.91}$\\
LF \cite{zeng2021rethinking} & $37.44$& $67.67$& $10.35$& $58.42$& $39.54$& $85.21$& $6.03$& $50.7$& $\underline{39.88}$& $\underline{28.83}$& $\underline{19.50}$& $\underline{79.06}$\\
SSBA \cite{li2021invisible} & $38.1$& $79.39$& $\underline{8.10}$& $54.48$& $\underline{40.65}$& $92.24$& $3.27$& $49.33$& $37.48$& $\textbf{0.31}$& $\textbf{11.27}$& $\textbf{93.71}$\\
WaNet \cite{nguyen2021wanet} & $37.54$& $59.23$& $\underline{15.21}$& $66.13$& $\underline{40.98}$& $96.09$& $2.21$& $49.42$& $38.50$& $\underline{1.44}$& $9.13$& $\textbf{95.50}$\\
Average & $39.26$& $66.65$& $12.47$& $60.38$& $\underline{41.66}$& $92.3$& $3.8$& $50.42$& $39.51$& $\underline{5.11}$& $\underline{14.47}$& $\textbf{86.86}$\\

\bottomrule
\end{tabular}}
\end{table}

\clearpage

\subsection{Learning curves for Shared Adversarial Risk}
\begin{figure}[h]
    \centering
    \includegraphics[width=\linewidth]{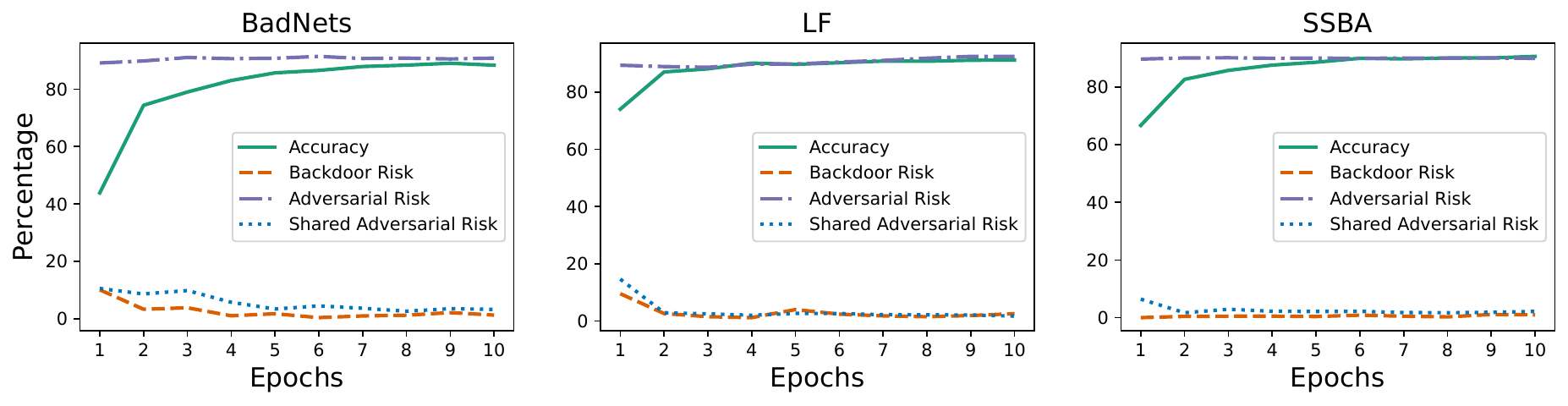}
    \caption{Learn curves for Accuracy, Backdoor Risk,  Adversarial Risk, and Shared Adversarial Risk.}
    \label{fig:asr}
\end{figure}

This section presents a plot of the learning curves for shared adversarial risk to illustrate the relationship between the backdoor risk (measured by ASR) and the shared adversarial risk. For each plot, we run SAU on CIFAR-10 with PreAct-ResNet18 and a poisoning ratio of 10\% for 10 epochs. The SAR is computed on the test dataset by computing the shared adversarial example using the inner maximization step in (8). Figure~\ref{fig:asr} demonstrates that SAU can effectively reduce the shared adversarial risk, and thus, mitigate the backdoor risk. We note that in order to satisfy Assumption 1, the perturbation set adopted by SAU is much larger than the perturbation budget in the adversarial robustness area, resulting in a high adversarial risk in Figure~\ref{fig:asr}.

\subsection{Experiments on different PGD configurations}
In this section, we test SAU using different PGD configurations to generate the shared adversarial examples. Specifically, we consider the following configurations for PGD:
\begin{itemize}
    \item Config 1: PGD with $L_\infty$ norm bound 0.2, step size $0.2$, and 5 steps.
    \item Config 2: PGD with $L_\infty$ norm bound 0.1, step size $0.1$, and 5 steps.
    \item Config 3: PGD with $L_\infty$ norm bound 0.2, step size $0.1$, and 5 steps.
    \item Config 4: PGD with $L_2$ norm bound 5, step size $0.2$, and 5 steps.
    \item Config 5: PGD with $L_1$ norm bound 300, step size $0.2$, and 5 steps.
\end{itemize}
The experiments are conducted for CIFAR-10 with PreAct-ResNet18 and poisoning ratio $10\%$. We summarize the experiment results in Table~\ref{pgd}, from which we can find the SAU can effectively mitigate backdoors under various configurations of PGD.
\begin{table}[H]
\centering
\caption{Results for SAU with different PGD  configurations}
\label{pgd}
\setlength{\tabcolsep}{3pt} 
\scalebox{0.72}{
\begin{tabular}{c|cccc|cccc|cccc}
\toprule
Defense & \multicolumn{4}{c|}{No Defense}       & \multicolumn{4}{c|}{Config 1 \textbf{(Default)}}& \multicolumn{4}{c}{Config 2}   \\ \midrule
Attack  & \multicolumn{1}{c}{ACC} & \multicolumn{1}{c}{ASR} & \multicolumn{1}{c}{R-ACC} & \multicolumn{1}{c|}{DER} & \multicolumn{1}{c}{ACC}         & \multicolumn{1}{c}{ASR}  & \multicolumn{1}{c}{R-ACC}& \multicolumn{1}{c|}{DER}  & \multicolumn{1}{c}{ACC}  & \multicolumn{1}{c}{ASR} & \multicolumn{1}{c}{R-ACC}& \multicolumn{1}{c}{DER}  \\
\midrule
BadNets \cite{gu2019badnets} & $91.32$& $95.03$& $4.67$&  N/A & $89.31$& $1.53$& $88.81$& $95.74$& $89.56$& $\textbf{0.80}$& $\textbf{89.60}$& $\textbf{96.24}$\\
Blended \cite{chen2017targeted} & $93.47$& $99.92$& $0.08$&  N/A & $90.96$& $6.14$& $\textbf{64.89}$& $95.63$& $91.05$& $\underline{6.06}$& $61.11$& $\underline{95.72}$\\
Input-Aware & $90.67$& $98.26$& $1.66$&  N/A & $\textbf{91.59}$& $1.27$& $\textbf{88.54}$& $98.49$& $90.93$& $1.56$& $84.76$& $98.35$\\
LF \cite{zeng2021rethinking} & $93.19$& $99.28$& $0.71$&  N/A & $90.32$& $4.18$& $\textbf{81.54}$& $96.12$& $90.83$& $3.76$& $77.93$& $\underline{96.58}$\\
SIG \cite{barni2019new} & $84.48$& $98.27$& $1.72$&  N/A & $\textbf{88.56}$& $1.67$& $\textbf{57.96}$& $98.30$& $\underline{88.29}$& $\underline{0.94}$& $49.41$& $\underline{98.66}$\\
SSBA \cite{li2021invisible} & $92.88$& $97.86$& $1.99$&  N/A & $\textbf{90.84}$& $1.79$& $\textbf{85.83}$& $\textbf{97.01}$& $89.81$& $\underline{1.63}$& $80.29$& $96.58$\\
WaNet \cite{nguyen2021wanet} & $91.25$& $89.73$& $9.76$&  N/A & $\textbf{91.26}$& $\textbf{1.02}$& $\textbf{90.28}$& $\textbf{94.36}$& $86.22$& $3.12$& $86.89$& $90.79$\\
Average & $91.04$& $96.91$& $2.94$&  N/A & $\textbf{90.41}$& $\underline{2.51}$& $\textbf{79.69}$& $\textbf{96.52}$& $89.53$& $2.55$& $75.71$& $96.13$\\

\toprule

\toprule
Defense & \multicolumn{4}{c|}{Config 3}  &  \multicolumn{4}{c|}{Config 4}        & \multicolumn{4}{c}{Config 5} \\
\midrule
Attack  & \multicolumn{1}{c}{ACC} & \multicolumn{1}{c}{ASR} & \multicolumn{1}{c}{R-ACC} & \multicolumn{1}{c|}{DER} & \multicolumn{1}{c}{ACC}         & \multicolumn{1}{c}{ASR}  & \multicolumn{1}{c}{R-ACC}& \multicolumn{1}{c|}{DER}  & \multicolumn{1}{c}{ACC}  & \multicolumn{1}{c}{ASR} & \multicolumn{1}{c}{R-ACC}& \multicolumn{1}{c}{DER}  \\
\midrule
BadNets \cite{gu2019badnets} & $\underline{89.69}$& $2.21$& $89.19$& $95.60$& $\textbf{89.78}$& $\underline{1.11}$& $\underline{89.38}$& $\underline{96.19}$& $89.04$& $1.29$& $88.63$& $95.73$\\
Blended \cite{chen2017targeted} & $90.98$& $6.12$& $61.64$& $95.65$& $\textbf{91.58}$& $8.31$& $59.09$& $94.86$& $\underline{91.49}$& $\textbf{5.83}$& $\underline{63.04}$& $\textbf{96.05}$\\
Input-Aware & $90.75$& $1.86$& $86.21$& $98.20$& $\underline{91.30}$& $\textbf{0.94}$& $84.24$& $\textbf{98.66}$& $91.21$& $\underline{1.20}$& $\underline{86.60}$& $\underline{98.53}$\\
LF \cite{zeng2021rethinking} & $\underline{91.05}$& $\underline{3.00}$& $\underline{81.23}$& $\textbf{97.07}$& $87.06$& $\textbf{1.94}$& $73.72$& $95.60$& $\textbf{91.29}$& $5.12$& $80.26$& $96.13$\\
SIG \cite{barni2019new} & $87.18$& $1.32$& $55.12$& $98.47$& $88.20$& $\textbf{0.78}$& $49.09$& $\textbf{98.74}$& $86.69$& $2.08$& $\underline{55.64}$& $98.09$\\
SSBA \cite{li2021invisible} & $90.39$& $2.68$& $82.04$& $96.34$& $\underline{90.65}$& $2.59$& $\underline{84.40}$& $96.52$& $90.48$& $\textbf{1.62}$& $84.19$& $\underline{96.92}$\\
WaNet \cite{nguyen2021wanet} & $\underline{91.14}$& $1.60$& $87.96$& $94.01$& $90.79$& $\underline{1.13}$& $\underline{89.70}$& $\underline{94.07}$& $90.55$& $1.82$& $89.11$& $93.61$\\
Average & $\underline{90.17}$& $2.68$& $77.63$& $\underline{96.48}$& $89.91$& $\textbf{2.40}$& $75.66$& $96.38$& $90.11$& $2.71$& $\underline{78.21}$& $96.44$\\

\bottomrule
\end{tabular}}
\end{table}

\subsection{Experiments on different numbers of clean samples}
This section examines the influence of clean sample size for SAU by evaluating SAU with different numbers of clean samples. The experiment is conducted on CIFAR-10 with PreAct-ResNet18 and the results are summarized in Table~\ref{clean_ratio}. Table~\ref{clean_ratio} shows that SAU can consistently mitigate backdoors with sample sizes ranging from 2500 (5\%) to 500 (1\%) with high accuracy. When the sample size decreases to 50 (0.1\%, 5 samples per class) or 10 (0.02\%, 1 sample per class), SAU can still reduce the backdoor to a low ASR. However, clean accuracy is difficult to guarantee with such limited clean samples.

\begin{table}[h]
\centering
\caption{Results for SAU with different numbers of clean samples}
\label{clean_ratio}
\setlength{\tabcolsep}{3pt} 
\scalebox{0.72}{
\begin{tabular}{@{}c|cc|cc|cc|cc|cc|cc@{}}
\toprule
Defense                      & \multicolumn{2}{c|}{No Defense}     & \multicolumn{2}{c|}{SAU (\textbf{Ours})-2500}     & \multicolumn{2}{c|}{SAU (\textbf{Ours})-1000}     & \multicolumn{2}{c|}{SAU (\textbf{Ours})-500}      & \multicolumn{2}{c|}{SAU (\textbf{Ours})-50}        & \multicolumn{2}{c}{SAU (\textbf{Ours})-10} \\ \midrule
\multicolumn{1}{c|}{Attack}  & ACC   & \multicolumn{1}{c|}{ASR}   & ACC   & \multicolumn{1}{c|}{ASR}  & ACC   & \multicolumn{1}{c|}{ASR}  & ACC   & \multicolumn{1}{c|}{ASR}  & ACC   & \multicolumn{1}{c|}{ASR}   & ACC          & ASR          \\ \midrule
\multicolumn{1}{c|}{BadNets \cite{gu2019badnets}} & $91.32$ & \multicolumn{1}{c|}{$95.03$} & $90.24$ & \multicolumn{1}{c|}{$1.30$} & $89.24$ & \multicolumn{1}{c|}{$1.31$} & $83.04$ & \multicolumn{1}{c|}{$2.17$} & $67.86$ & \multicolumn{1}{c|}{$1.41$}  & $51.43$        & $14.38$        \\ \midrule
\multicolumn{1}{c|}{Blended \cite{chen2017targeted}} & $93.47$ & \multicolumn{1}{c|}{$99.92$} & $91.09$ & \multicolumn{1}{c|}{$3.83$} & $88.82$ & \multicolumn{1}{c|}{$2.53$} & $87.71$ & \multicolumn{1}{c|}{$5.41$} & $73.95$ & \multicolumn{1}{c|}{$11.62$} & $43.54$        & $0.00$         \\ \midrule
LF \cite{zeng2021rethinking}                           & $93.19$ & $99.28$                      & $87.19$ & $2.98$                      & $88.00$ & $6.79$                      & $88.69$ & $2.84$                      & $66.69$ & $0.30$                       & $54.95$        & $12.62$        \\ \bottomrule
\end{tabular}}
\end{table}

\subsection{Experiments on ALL to ALL attack}
In this section, we compare SAU with other baselines on ALL to ALL attacks on CIFAR-10 with PreAct-ResNet18 and poisoning ratio $10\%$.  Specifically, the target labels for the sample with original labels $y$ are set to $y_t=(y+1) \mod{K}$ where $\mod$ is short for "modulus". The experiment results are summarized in Table~\ref{cifar10_preactresnet18_a2a}. From Table~\ref{cifar10_preactresnet18_a2a}, we can find that SAU achieves the best defending performance in 5 of 7 attacks and the lowest average ASR. At the same time, SAU also achieves the best average R-ACC and average DER, which further demonstrates its effectiveness in defending against backdoor attacks with multiple targets.

Note that the ASR for the poisoned model is significantly lower than the ASR in the single-target case. Thus, Table~\ref{cifar10_preactresnet18_a2a} also shows that SAU can still effectively mitigate backdoor even the ASR of the poisoned model is much lower than $100\%$.
\begin{table}[h]
\centering
\caption{Results for defending against ALL to ALL attacks}
\label{cifar10_preactresnet18_a2a}
\setlength{\tabcolsep}{3pt} 
\scalebox{0.68}{
\begin{tabular}{c|cccc|cccc|cccc|cccc}
\toprule
Defense & \multicolumn{4}{c|}{No Defense}    & \multicolumn{4}{c|}{ANP \cite{wu2021adversarial}}    & \multicolumn{4}{c|}{FP \cite{liu2018fine}}& \multicolumn{4}{c}{NC \cite{wang2019neural}}   \\ \midrule
Attack  & \multicolumn{1}{c}{ACC} & \multicolumn{1}{c}{ASR} & \multicolumn{1}{c}{R-ACC} & \multicolumn{1}{c|}{DER}& \multicolumn{1}{c}{ACC}  & \multicolumn{1}{c}{ASR} & \multicolumn{1}{c}{R-ACC} & \multicolumn{1}{c|}{DER}  & \multicolumn{1}{c}{ACC}         & \multicolumn{1}{c}{ASR}  & \multicolumn{1}{c}{R-ACC}& \multicolumn{1}{c|}{DER}  & \multicolumn{1}{c}{ACC}  & \multicolumn{1}{c}{ASR} & \multicolumn{1}{c}{R-ACC}& \multicolumn{1}{c}{DER}  \\
\midrule
BadNets \cite{gu2019badnets} & $91.89$& $74.42$& $18.66$&  N/A & $\textbf{92.33}$& $2.56$& $\underline{90.12}$& $\underline{85.93}$& $83.91$& $1.72$& $83.71$& $82.36$& $\underline{91.88}$& $74.41$& $18.67$& $50.0$\\
Blended \cite{chen2017targeted} & $93.67$& $86.69$& $2.0$&  N/A & $\underline{93.25}$& $51.35$& $36.39$& $67.46$& $84.43$& $\underline{3.53}$& $69.05$& $86.96$& $\textbf{93.67}$& $86.69$& $2.0$& $50.0$\\
Input-Aware & $91.92$& $83.69$& $7.23$&  N/A & $91.04$& $\textbf{0.85}$& $\underline{88.34}$& $\textbf{90.98}$& $85.69$& $2.01$& $81.09$& $87.72$& $\underline{91.20}$& $81.49$& $8.75$& $50.74$\\
LF \cite{zeng2021rethinking} & $93.84$& $89.96$& $2.44$&  N/A & $\underline{93.10}$& $2.58$& $\textbf{90.31}$& $\underline{93.32}$& $84.54$& $\underline{1.77}$& $80.8$& $89.44$& $\textbf{93.84}$& $89.96$& $2.44$& $50.0$\\
SIG \cite{barni2019new} & $93.4$& $90.61$& $1.09$&  N/A & $\underline{92.83}$& $86.56$& $2.69$& $51.74$& $85.01$& $6.09$& $59.7$& $88.06$& $\textbf{93.40}$& $90.61$& $1.09$& $50.0$\\
SSBA \cite{li2021invisible} & $93.46$& $87.84$& $3.7$&  N/A & $\underline{93.29}$& $7.10$& $\underline{84.36}$& $\underline{90.29}$& $82.97$& $\underline{2.48}$& $80.26$& $87.44$& $\textbf{93.46}$& $87.84$& $3.7$& $50.0$\\
WaNet \cite{nguyen2021wanet} & $89.91$& $78.58$& $10.62$&  N/A & $90.27$& $\textbf{1.00}$& $88.91$& $\textbf{88.79}$& $86.62$& $1.65$& $84.55$& $86.82$& $89.91$& $78.58$& $10.61$& $50.0$\\
Average & $92.58$& $84.54$& $6.53$&  N/A & $\underline{92.30}$& $21.71$& $68.73$& $81.22$& $84.74$& $\underline{2.75}$& $77.02$& $86.97$& $\textbf{92.48}$& $84.23$& $6.75$& $50.11$\\

\toprule

\toprule
Defense & \multicolumn{4}{c|}{NAD \cite{li2021neural}}  & \multicolumn{4}{c|}{EP \cite{zheng2022preactivation}} & \multicolumn{4}{c|}{i-BAU \cite{zeng2022adversarial}}        & \multicolumn{4}{c}{SAU (\textbf{Ours})} \\
\midrule
Attack  & \multicolumn{1}{c}{ACC} & \multicolumn{1}{c}{ASR} & \multicolumn{1}{c}{R-ACC} & \multicolumn{1}{c|}{DER}  &  \multicolumn{1}{c}{ACC} & \multicolumn{1}{c}{ASR} & \multicolumn{1}{c}{R-ACC} & \multicolumn{1}{c|}{DER}  & \multicolumn{1}{c}{ACC}         & \multicolumn{1}{c}{ASR}  & \multicolumn{1}{c}{R-ACC}& \multicolumn{1}{c|}{DER}  & \multicolumn{1}{c}{ACC}  & \multicolumn{1}{c}{ASR} & \multicolumn{1}{c}{R-ACC}& \multicolumn{1}{c}{DER}  \\
\midrule
BadNets \cite{gu2019badnets} & $80.49$& $2.51$& $80.54$& $80.25$& $88.72$& $3.00$& $87.94$& $84.12$& $89.39$& $\underline{1.29}$& $89.90$& $85.32$& $90.69$& $\textbf{1.02}$& $\textbf{90.61}$& $\textbf{86.10}$\\
Blended \cite{chen2017targeted} & $76.85$& $5.63$& $58.85$& $82.12$& $91.69$& $82.62$& $3.66$& $51.05$& $90.46$& $8.83$& $\underline{72.20}$& $\underline{87.32}$& $91.47$& $\textbf{2.17}$& $\textbf{82.08}$& $\textbf{91.16}$\\
Input-Aware & $82.54$& $1.75$& $76.86$& $86.28$& $90.2$& $\underline{1.13}$& $87.69$& $90.42$& $89.40$& $6.03$& $81.10$& $87.57$& $\textbf{91.74}$& $1.87$& $\textbf{89.12}$& $\underline{90.82}$\\
LF \cite{zeng2021rethinking} & $73.58$& $2.83$& $74.55$& $83.43$& $91.88$& $84.64$& $5.97$& $51.68$& $89.38$& $7.88$& $64.68$& $88.81$& $91.67$& $\textbf{1.11}$& $\underline{85.84}$& $\textbf{93.34}$\\
SIG \cite{barni2019new} & $79.99$& $5.78$& $55.48$& $85.71$& $90.91$& $86.10$& $2.44$& $51.01$& $90.74$& $\underline{1.57}$& $\textbf{86.31}$& $\underline{93.19}$& $91.13$& $\textbf{1.48}$& $\underline{85.14}$& $\textbf{93.43}$\\
SSBA \cite{li2021invisible} & $74.12$& $3.58$& $69.61$& $82.46$& $90.33$& $41.83$& $46.59$& $71.44$& $88.56$& $6.49$& $78.49$& $88.22$& $91.35$& $\textbf{1.84}$& $\textbf{85.53}$& $\textbf{91.94}$\\
WaNet \cite{nguyen2021wanet} & $85.26$& $1.99$& $82.59$& $85.97$& $86.66$& $74.74$& $11.13$& $50.3$& $\underline{91.71}$& $1.63$& $\textbf{90.98}$& $88.48$& $\textbf{91.73}$& $\underline{1.43}$& $\underline{89.34}$& $\underline{88.58}$\\
Average & $78.98$& $3.44$& $71.21$& $83.75$& $90.06$& $53.44$& $35.06$& $64.29$& $89.95$& $4.82$& $\underline{80.52}$& $\underline{88.42}$& $91.40$& $\textbf{1.56}$& $\textbf{86.81}$& $\textbf{90.77}$\\

\bottomrule
\end{tabular}}
\end{table}





\end{document}